\documentclass[11pt,twocolumn,twoside]{article}
\usepackage{iccr2024}
\usepackage{color,soul}
\usepackage{cleveref}
\usepackage{pdflscape}
\usepackage{xspace}
\newcommand{\ie}{i.e.,\xspace{}}

\usepackage{float}

\begin{document}

\title{Replication Study and Benchmarking of\\Real-Time Object Detection Models} 

\author[1]{Pierre-Luc Asselin}
\author[2]{Vincent Coulombe}
\author[3]{William Guimont-Martin}
\author[3]{William Larrivée-Hardy}

\affil[1]{Département de physique, de génie physique et d’optique,
        Université Laval, Québec, Québec, Canada.}
\affil[2]{Département de génie électrique et de génie informatique,
        Université Laval, Québec, Québec, Canada.}
\affil[3]{Département d'informatique et de génie logiciel,
        Université Laval, Québec, Québec, Canada.}
\affil[*]{Authors are presented in alphabetical order, each having equal contribution to the work.}
\affil[**]{This report was written in the context of the course GIF-7010 Avancées en apprentissage automatique at Université Laval.}

\maketitle
\thispagestyle{fancy}

\section*{Reproducibility summary}

This work examines the reproducibility and benchmarking of state-of-the-art real-time object detection models. As object detection models are often used in real-world contexts, such as robotics, where inference time is paramount, simply measuring models' accuracy is not enough to compare them.
We thus compare a large variety of object detection models' accuracy and inference speed on multiple graphics cards.
In addition to this large benchmarking attempt, we also reproduce the following models from scratch using PyTorch on the MS COCO 2017 dataset: DETR, RTMDet, ViTDet and YOLOv7. More importantly, we propose a unified training and evaluation pipeline, based on MMDetection's features, to better compare models.
\newline
\newline
Our implementation of DETR and ViTDet could not achieve accuracy or speed performances comparable to what is declared in the original papers.
On the other hand, reproduced RTMDet and YOLOv7 could match such performances.
Studied papers are also found to be generally lacking for reproducibility purposes.
As for MMDetection pretrained models, speed performances are severely reduced with limited computing resources (larger, more accurate models even more so).
Moreover, results exhibit a strong trade-off between accuracy and speed, prevailed by anchor-free models --- notably RTMDet or YOLOx models.
\newline
\newline
The code used in this paper and for all the experiments is available in the repository at \url{https://github.com/willGuimont/segdet_mlcr2024}.

\section{Introduction}

Computer vision, as a machine learning task, has seen unprecedented growth in the last ten years.
In 2012, AlexNet showed the proficiency of large convolutional neural networks (CNN) for object detection, leading to increasingly larger and more accurate state-of-the-art CNNs.
These models offer great accuracy, but often require significant inference time to obtain predictions.
In 2014, R-CNN opened the way to real-time object detection models by limiting inference time while still maintaining competitive accuracy~\cite{R-CNN}.
Since then, different architectures and benchmarks have emerged.
This paper wishes to build upon current real-time object detection benchmark methods and reproduce state-of-the-art models.

Object Detection models locate objects of interest in an image or video by identifying the position and boundaries of objects, and classifying them into different categories.
As such, typical object detection architecture is divided into three main parts.
The \textit{backbone} is used as a feature extractor, giving a feature map representation of the input image.
Then, the backbone feeds the feature map to a \textit{neck} which aggregates and produces higher-level features (typically with a feature pyramid network)~\cite{FPN}.
The object detection task itself is performed by the \textit{head} on the feature map produced by the backbone and the neck.
\newline
\newline
When evaluating object detection models, it's tempting to rely solely on metrics like mean Average Precision (mAP) to gauge their performance.
While mAP provides a useful summary of a model's precision across different object categories, it's essential to recognize its limitations and consider additional factors in model assessment.
mAP focuses solely on detection accuracy and overlooks other crucial aspects such as model speed, resource efficiency, and robustness to real-world variations.
A high mAP doesn't necessarily translate to practical usability in some contexts, such as robotics, if the model is too computationally intensive.
\newline
\newline
This comparative study aims to reproduce a selection of models representative of current techniques in real-time object detection, more specifically YOLOv7~\cite{YOLOv7}, RTMDet~\cite{RTMDet}, ViTDet~\cite{ViTDet} and DETR~\cite{DETR}.
Moreover, an evaluation pipeline was developed in order to fairly compare the different models with each other on precision, frame rate and model size.
This pipeline was also applied to a larger selection of models that are already trained so that we can compare them together and draw conclusions.
\\
In order to better compare the different approaches taken for object detection, we offer a review of the three primary approaches as outlined in \cite{MaskBEV}.
Object detection methods mainly follow three approaches: anchor-based detectors, anchor-free detectors, and attention-based detectors.

\subsection*{Anchor-based Detection}

Anchor-based methods employ predefined detection boxes, called anchors, to predict object location and size~\cite{MaskBEV}.
Two anchor-based methods families are prominent.
Models branching from R-CNN are mainly two-stage detectors: first generating proposed regions with a selective search algorithm before doing the object classification task for the selected regions~\cite{R-CNN}.
Other models, such as YOLOv7, are instead single-stage detectors where object classification and bounding-box regression are done directly, without using pre-generated region proposals~\cite{YOLOv7}.
\newline
\newline
A greater quantity of anchors allows such models to better recognize objects of different shapes and sizes, leading to increased performances~\cite{CornerNet}.
However, anchor-based methods demand various hyperparameters to define the anchor box characteristics, burdening training costs and hyperparameter tuning, and thus representing a strong inductive biais.
Additionally, anchor-based performances are conditional on how well anchor boxes align with objects and may be limited with overlapping objects~\cite{Anchor-Quality}.

\subsection*{Anchor-free Detection}

Anchor-free detectors often utilize feature maps or specific pixel characteristics to predict object locations and sizes.
For example, CornerNet detects objects by predicting their geometrical center~\cite{CenterNet}.
This technique removes the need for predefined anchor boxes, thereby enabling the development of lighter and consequently faster models.
However, this method relies on more intricate post-processing techniques like thresholding, max pooling, and non-maximum suppression (NMS).
Nonetheless, anchor-free detectors require less post-processing time than anchor-based detectors using NMS post-processing and maintain equivalent accuracy~\cite{RT-DETR}.

\subsection*{Attention-based Detection}
Attention-based architectures, such as Transformers, are originally from the natural language processing (NLP) field but have more recently made breakthroughs in vision tasks~\cite{Attention}, due to the advantages of the attention mechanism.
Detectors based on transformers remove the need for anchors and feature maps by directly predicting bounding boxes and class labels using the cross-attention mechanism of the transformer with a set of learnt positional embeddings~\cite{DETR}.
The transformer encoder uses self-attention to extract information from the image, and the transformer decoder uses cross-attention between this extracted data and the learnt positional embeddings.
The decoder's output embedding is then used by a classification head and a feedforward network to predict class and bounding box location, respectively.
It is important to note that such models use a limited number of detection boxes fixed by the number of learnt positional embeddings.
As such, models based on this approach can forgo post-processing steps and directly predict unique bounding boxes~\cite{DETR}.

\section{Scope of reproducibility and benchmark}

No original author was contacted during this work.
Implementations were solely based on selected original papers~\cite{DETR, RTMDet, YOLOv7, ViTDet}.
If insufficient information is given in the paper, the official code repository was consulted.
Every instance where this was necessary is mentioned explicitly. 
Additionally, no pretraining method was reproduced.
Pretrained weights for the backbones were used when available.
\newline
\newline
The additional benchmarked models were taken from MMDetection's catalogue of pretrained models.
Due to time constraints, we selected an array of models representing state-of-the-art methods.
We used these models as provided and ran them in our evaluation pipeline.

\section{Methodology}

\subsection{Metrics}

The usability of deep neural networks is not only based on accuracy, but also on model size, inference speed and post-processing load required to use the output.
To better capture a model's practicality, the following metrics are measured:

\begin{itemize}
    \setlength\itemsep{0em}
    \item Mean average precision from 0.5 IoU to 0.95 with 0.05 steps (mAP);
    \item Average precision at 0.5 IoU (AP50);
    \item Average precision at 0.75 IoU (AP75);
    \item Average precision for small objects (APs);
    \item Average precision for medium objects (APm);
    \item Average precision for large objects (APl);
    \item Network size;
    \item Prediction time for a batch of 1 image (including all required post-processing);
    \item Prediction time for a batch of 16 image (including all required post-processing); and
    \item Prediction time for a batch of 32 image (including all required post-processing).
\end{itemize}

The recorded inference time considers the entire process from input processing to obtaining usable output, ensuring representability of real-world applications.
Studying the prediction time across multiple batch sizes is crucial to understand how a model's performances scales.
This is done by estimating amortized inference time, \ie{} the average time taken for a batch.
This metric could be representative of \textit{almost real-time} tasks where detection need to be done on a small batch of buffered inputs --- \ie{} scenarios where timely processing of a few consecutive frames is essential.

\subsection{Experimental setup and code}
\label{sec:method:exp}

\begin{table}[ht]
    \centering
    \caption{Hardware used for our study.}
    \begin{tabular}{lllr}
    \toprule
    \textbf{Nb} & \textbf{GPU} & \textbf{CPU} & \textbf{RAM} \\
    \midrule
    1 & Titan X & AMD Ryzen 7 1700 & 16 Gb \\
    1 & Quadro RTX-8000 & i7-6700K & 32 Gb \\
    4 & RTX A6000 & AMD Ryzen 3970X & 128 Gb \\
    1 & RTX-3080 Ti & i7-12700K & 64 Gb \\
    1 & RTX-4090 & i9-13900F & 64 Gb \\
    \bottomrule
    \end{tabular}
    \label{tab:hardware}
\end{table}

Model training was exclusively made using computing resources at the Northern Robotics Laboratory (NorLab) from Université Laval.
GPUs (see table \ref{tab:hardware}) were selected in such a way as to offer a range of performance, memory and release date.
Model training was made exclusively with the 4 available A6000 GPUs without any multi-GPU training.
Training was made within a Docker environment,\footnote{Based on docker image\\ \texttt{pytorch/pytorch:2.2.1-cuda12.1-cudnn8-devel}} using Slurm\footnote{\url{https://slurm.schedmd.com/overview.html} consulted April 19th 2024.} for job scheduling.
This contributes to standardizing the software stacks while ensuring the possibility to port both training and evaluation to various machines.
A unified pipeline simplifies model comparisons on different hardware.
\newline
\newline
All models listed in section \ref{sec:method:models} were evaluated through this pipeline.
Before each evaluation, a few iterations are made to avoid cold starts.
First, the precision metrics are evaluated on the validation set of MS COCO 2017 (see \ref{sec:method:dataset} for more details), followed by a measurement of the total prediction time on all images.
This inference time estimation includes any post-processing used by the model, non-maximum suppression for instance.
This is done on batch sizes of 1, 16 and 32.
\newline
\newline
All used code for this reproducibility project is available in the repository at \url{https://github.com/willGuimont/segdet_mlcr2024} (April 21st 2024).

\subsection{Datasets}
\label{sec:method:dataset}

For all training and testing, only MS COCO 2017,\footnote{\url{https://cocodataset.org/\#home} consulted April 19th 2024.} was used.
This dataset contains 118K images in its training set and 5K images for validation.
All images have bounding boxes and masks over 80 object categories. 
Reported average precisions are computed using the python library \texttt{pycocotools}.\footnote{\url{https://pypi.org/project/pycoco/} consulted April 19th 2024.}

\subsection{Models}
\label{sec:method:models}

Four models in real-time object detection were selected for reproduction, based on their performances, but also to be representative of current state-of-the-art approaches.

\subsubsection*{Yolov7}

First introduced in 2015, \textit{You Only Look Once} is a family of real-time object detection algorithms that stands out for its balance of accuracy and speed~\cite{YOLO}.
The YOLO architecture unifies object detection by trying to detect every bounding box simultaneously, framing the problem as a regression problem instead of a classification problem (by spatially separating bounding boxes and predicting their respective position and probability).
This allows YOLO models to give predictions in a single pass of the network. 
Following iterations of YOLO gradually boosted the architecture's performances. 
Since YOLOv2, anchor-based detection is used by this family to further improve accuracy (with the notable exclusion of YOLOx~\cite{YOLOX} models)~\cite{YOLOv2}.
At the time of its release in 2022, YOLOv7 surpassed every other object detection model in both accuracy and speed (for the range of 5 to 160 FPS)~\cite{YOLOv7}.
YOLOv7 is largely based on YOLOv5~\cite{YOLOv5}, but incorporates \textit{implicit knowledge} from YOLOR~\cite{YOLOR}.
Implicit knowledge is similar to human past experience knowledge.
YOLOv7 therefore uses information on prior tasks for its predictions.
\newline
\newline
YOLOV7 also includes a \textit{bag-of-freebies}: a series of optional features that increase the model's accuracy without affecting inference time.
These include: a planned re-parametrization of convolution blocks, coarse label assignment in an auxiliary head during training, batch normalization, exponential moving average for the final inference, etc.
However, these components do affect training time, as noted by the original authors~\cite{YOLOv7}.
The code behind the model is open-source and comes with instructions for out-of-the-box usage with MS COCO 2017.\footnote{\url{https://github.com/WongKinYiu/yolov7}, consulted April 19th 2024.}

\subsubsection*{RTMDet}

RTMDET (\textit{Real-Time Models for object DETection}) is a single-stage, anchor-free approach to object detection~\cite{RTMDet}.
It leverages innovative architectural features, including wide depth-wise convolution to enhance performances of YOLOx's architecture~\cite{YOLOX}.
Moreover, RTMDet employs advanced training optimizations such as: diverse augmentation strategies, exponential moving average and dynamic soft label assignments.
\newline
\newline
A CSPNeXt backbone is introduced to maximize the effective receptive field while ensuring computational efficiency. 
CSPNeXt is a stack of wide and shallow CSPNeXt blocks. 
Each block exploits a 5x5 depth-wise convolution, which is identified as the optimal kernel size to expand the receptive field without significant computational overhead~\cite{RTMDet}. 
Additionally, CSPNeXt integrates channel-wise attention mechanisms after each stage to reduce the total number of blocks (24 to 18) while maintaining performances.
\newline
\newline
A multiscale feature pyramid network acts as neck~\cite{FPN}, mirroring the same architectural strategy as the backbone. 
Due to the relatively shallow nature of each feature pyramid scale level, no skip connections are used between its blocks.
It is worth noting that the neck’s total parameter count is similar to the backbone’s, as empirical evidence suggests that backbone-neck symmetry yields better performances~\cite{RTMDet}. 
\newline
\newline
The model uses a shared detection head for all scales, which reduces parameter count by approximately 10\% without impeding performances~\cite{RTMDet}.
However, each scale within the pyramid does not share the same batch normalization layer, as the statistical characteristics of features differ for each scale~\cite{RTMDet}. 
\newline
\newline
The loss function (equation \ref{eq:loss_rtmdet}) is composed of a Quality Focal Loss (QFL)~\cite{GFL} and a General Intersection Over Union (GIoU)~\cite{GIOU}.
The labels are assigned via a novel dynamic soft label assignment strategy, leveraging SimOTA~\cite{YOLOX} methodology. 

\begin{equation}\label{eq:loss_rtmdet}
\mathcal{L} = \lambda_1 \cdot \text{GIoU} + \lambda_2 \cdot \text{QFL}
\end{equation}

where \( \lambda_1 = 2 \) and \( \lambda_2 = 1 \).
\newline
\newline
The training pipeline uses a similar augmentation strategy as YOLOX such as the usage of MixUp~\cite{MIXUP} and Mosaic~\cite{YOLOv3} for approximately the initial 90\% of the training duration.
However, the authors avoided random rotation and shearing, as they cause misalignment between box annotations and the inputs.
Regarding optimization, AdamW~\cite{ADAM} is preferred over SGD due to empirical evidence indicating unstable convergence progress with the latter~\cite{RTMDet}, especially when subjected to heavy data augmentation during training.
The learning rate starts with a warm-up and remains constant until halfway through training. Subsequently, it transitions to cosine annealing for the latter half.

\subsubsection*{ViTDet}

Typically, object detection heads rely on multiscale feature maps produced from a hierarchical backbone --- such is the case for models using ResNet~\cite{RESNET} or Swin~\cite{Swin} backbones.
Vision Transformers have shown to be powerful visual recognition backbones in image classification, but are rarely used directly in object detection since they are plain and non-hierarchical, and thus, not adapted to most object detection heads~\cite{ViT}.
VitDet proposes the use of Simple Feature Pyramids (SFP) as necks to generate multiscale feature maps that can adapt any backbone network to any detection head --- including non-hierarchical backbones.
SFP uses a series of simple convolution and deconvolution layers to generate maps of different scales ($\{\frac{1}{32}, \frac{1}{16}, \frac{1}{8}, \frac{1}{4}\}$) using only the single-scale output from a ViT.
\newline
\newline
VitDet uses a ViT plain backbone directly with a Cascade Mask R-CNN head.
Performances on COCO datasets are competitive with other architectures based on ViT (but adapting it for hierarchical structures) such as Swin, while having faster inference and training times~\cite{ViTDet}.
ViTDet has opened the way to general-purpose backbones, with minimal architecture changes for different tasks.
The code used by the original authors is completely open-source.

\subsubsection*{DETR}
DETR (\emph{DEtection TRansformer}) is a transformer-based architecture composed of a CNN backbone, an encoder-decoder pair as a neck and a feed-forward network (FFN) head.
Images undergo processing within the CNN backbone, generating a lower-resolution activation map.
The typical output dimensions are $C = 2048$ channels of spatial dimensions $(H, W) = (\frac{H_0}{32}, \frac{W_0}{32})$, where $(H_0, W_0)$ is the input's size.
This activation map is then passed through a 1x1 convolution to reduce channel dimension and split into a sequence of patches, before reaching the encoder.
Each encoder layer consists of a multi-head self-attention block followed by a FFN.
Positional encodings are added to the input at each attention layer to preserve spatial information.
\newline
\newline
From the input sequence, the decoder generates $N$ output embeddings.
These output embeddings make up the input of a 3-layer perceptron with ReLU activation acting as the architecture's head.
This FFN predicts the normalized coordinates and dimensions of the bounding box, and a linear projection layer predicts the class using a softmax function.
A special class label named \textit{No object} is added to represent when a detection does not correspond to an object.
This addition is required since the number of predicted boxes is fixed, depending on the learned embedding fed into the decoder.
Predicted boxes and labels are assigned to ground truth using Hungarian matching.
Finally, the loss is computed by adding an IoU bounding box loss to a classification loss.

\subsubsection*{Other methods}

The open-source MMdetection toolbox\footnote{\url{https://github.com/open-mmlab/mmdetection}, consulted April 21st 2024.} (offered by the OpenMMLab project) provides different pretrained model implementations.
Since MMDetection is at the core of our training pipeline, we also considered performances for the following pretrained models:
Mask R-CNN R50, Mask R-CNN R101, Mask R-CNN X101, Mask R-CNN Swin, Mask R-CNN Swin (Crop)~\cite{he2017mask},
Faster R-CNN R50, Faster R-CNN Swin, Faster R-CNN X101~\cite{girshick2015fast},
DETR, DETR R50~\cite{DETR},
Deformable DETR R50 (2 stages), Deformable DETR R50~\cite{zhu2020deformable}, 
Cond. DETR R50~\cite{meng2021conditional},
DAB DETR R50~\cite{liu2022dab},
YOLOv3, YOLOv3 d53~\cite{YOLOv3}, 
MobileNetV2~\cite{sandler2018mobilenetv2}, 
DINO-5 Swin, DINO-4 R50~\cite{zhang2022dino},
YOLOf R50~\cite{chen2021you}, 
CenterNet R50~\cite{CenterNet}, 
RetinaNet~\cite{lin2017focal}, 
YOLOX-tiny, YOLOX-s, YOLOX-l, YOLOX-x~\cite{YOLOX}, 
RTMDet-ins tiny, RTMDet-ins m, RTMDet-x and RTMDet~\cite{RTMDet}.

\subsection{Implementation details}

Model training code uses MMDetection, an open-source object detection toolbox.
The rest of this section give additional implementation information for each reproduced model.

\subsubsection*{Yolov7}
The Yolov7 base model was reimplemented using PyTorch modules.
We did not use any pretrained weights from the original authors.
Additionally, Yolov7's custom data loader, custom loss function and data augmentation were taken directly from the original code repository and not reimplemented from scratch (this required the implementation of a compatible argument parser).
No feature from Yolov7's \textit{bag-of-freebies} was used during training or testing.
Due to difficulties adapting YOLOv7's data loader to OpenMMLab's architecture, this reproduction is the only one not using MMDetection modules for training.
This issue is recognized by OpenMMLab, which provides another toolbox specifically for YOLO family models.\footnote{\url{https://github.com/open-mmlab/mmyolo}, consulted April 25th 2024} 
All hyperparameters used are presented in table \ref{tab:hyperparameters}. 
These hyperparameters match those used for Yolov7's base model.
Only the batch size is different, being 64 instead of 32 to accelerate training.

\subsubsection*{RTMDet}
The RTMDet small model was reproduced using the MMdetection~\cite{MMDETECTION} toolbox.
As such, we did not reimplement the dynamic label assignments with soft labels, the augmentation pipeline, the loss calculation and the post-processing steps including NMS~\cite{YOLO}, as they were already integrated within the framework.
We trained the model on MS COCO 2017 using the hyperparameters presented in Table \ref{tab:hyperparameters}.

\subsubsection*{ViTDet}
The ViTDet base model (ViT backbone with Cascade Mask R-CNN head) was reimplemented as described in the original paper using the MMDetection toolbox.
We implemented a custom neck implementing the SFP generating multiscale features from the output of a ViT backbone.
This backbone is a ViT pretrained with MAE~\cite{Masked}.
The output of the SFP is then fed into a CascadeRoIHead head from MMDetection.
While the original paper did not describe the use of any data augmentation pipeline, we opted to replicate the same data augmentation used in RTMDet.
All hyperparameters used are presented in table \ref{tab:hyperparameters}.
Only the batch size is different from the original paper, being 24 per GPU instead of 1 per GPU.

\subsubsection*{DETR}
The DETR base model was reproduced as described in the original paper~\cite{DETR}, the only difference being the batch size used during training (40 instead of 64).
As described in the original paper, the PyTorch implementation of Transformers cannot add positional encoding at the input of each layer.
Thus, we used the implementation of Transformer from \texttt{willGuimont/transformers}~\cite{WGMTransformer}.\footnote{\url{https://github.com/willGuimont/transformers}, consulted on April 28th 2024}
We used the standard pretrained ResNet50 ~\cite{RESNET} from PyTorch for the backbone.

\subsubsection*{Pretrained models}
For the other considered approaches, we automated the evaluation pipeline as much as possible.
First, we parse every model description present in the MMDetection GitHub repository to find all the models providing pretrained weights.
We then, due to time constraints, selected an array of methods to benchmark.
Our pipeline downloads automatically the configuration file, containing everything needed to reproduce the model in the MMDetection framework, and the corresponding weights.
We then run the models through our unified evaluation benchmark and collect the inference statistics.

\section{Results}

\begin{figure*}[ht]
\centering
     \begin{minipage}{0.3\textwidth}
         \includegraphics[width=\textwidth]{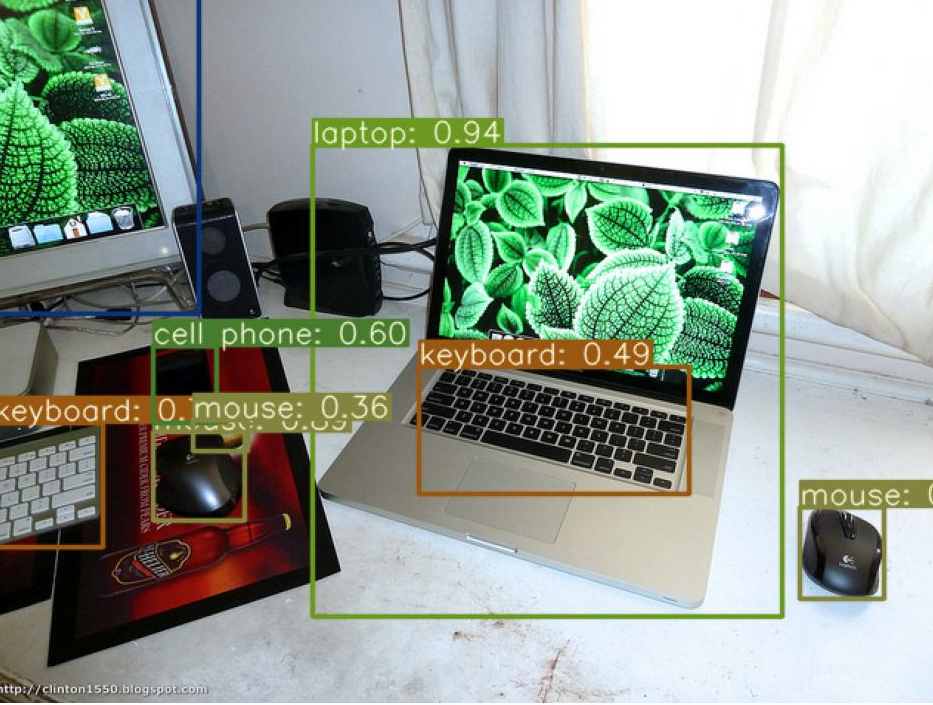}
     \end{minipage}
     \begin{minipage}{0.3\textwidth}
         \includegraphics[width=\textwidth, height=4.3cm]{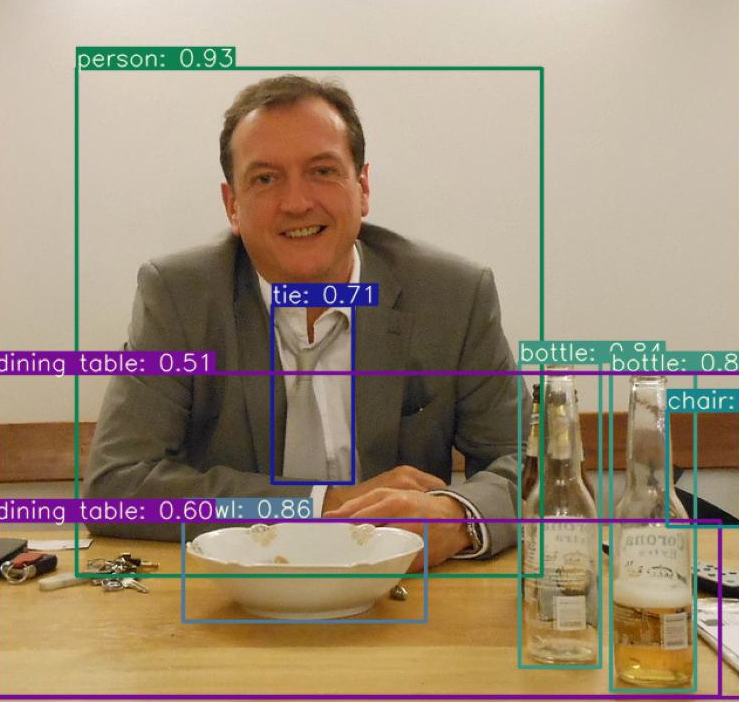}
     \end{minipage}
     \begin{minipage}{0.3\textwidth}
         \includegraphics[width=\textwidth]{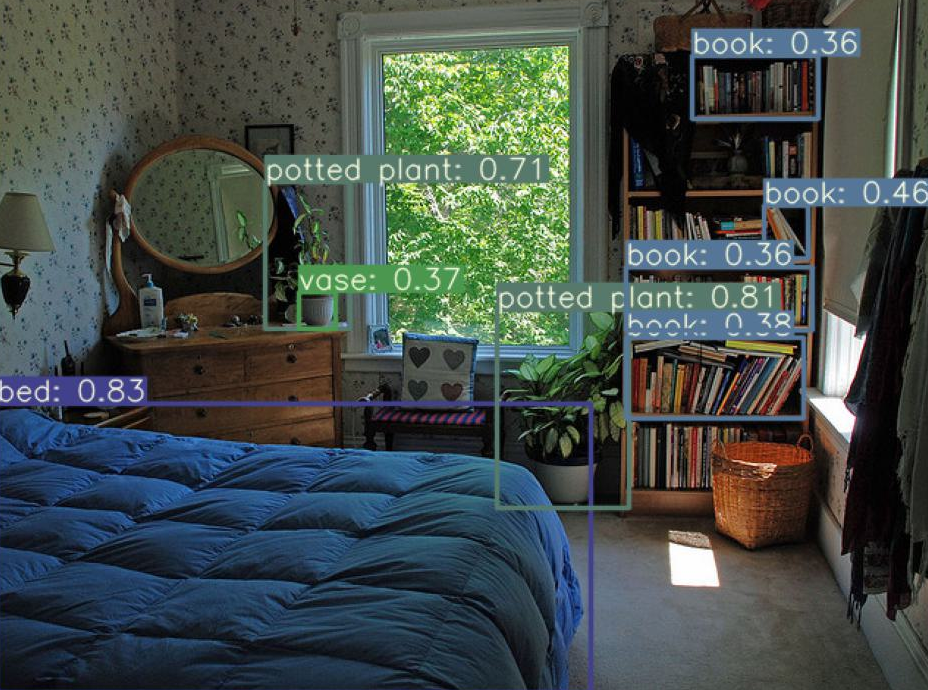}
     \end{minipage}

     \vskip\baselineskip
     \begin{minipage}{0.3\textwidth}
         \includegraphics[width=\textwidth]{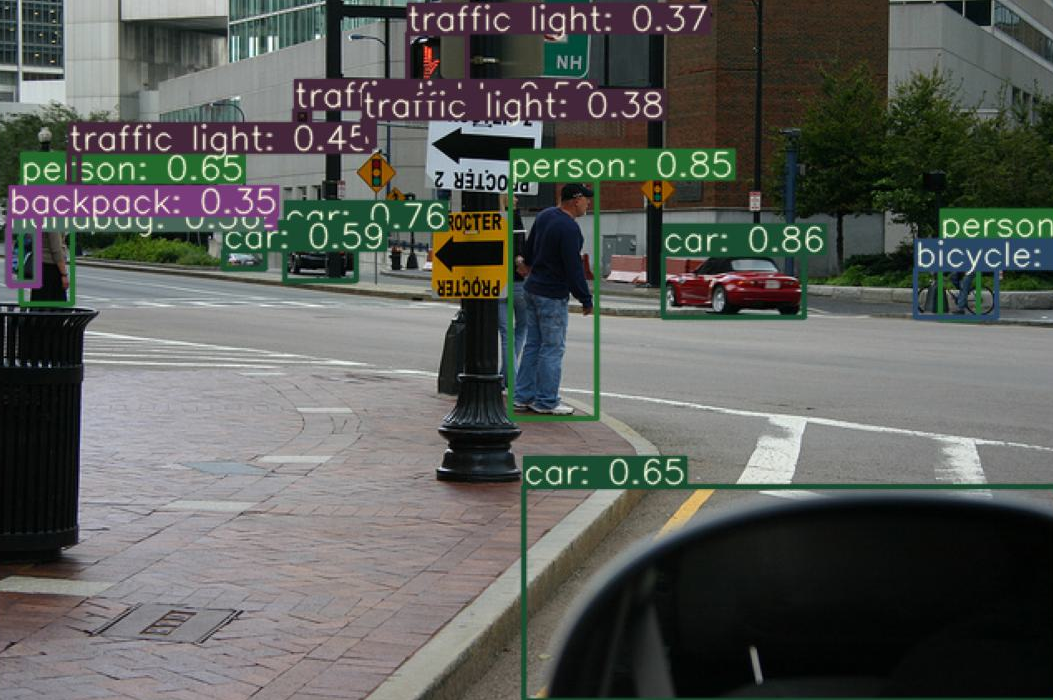}
     \end{minipage}
     \begin{minipage}{0.3\textwidth}
         \includegraphics[width=\textwidth]{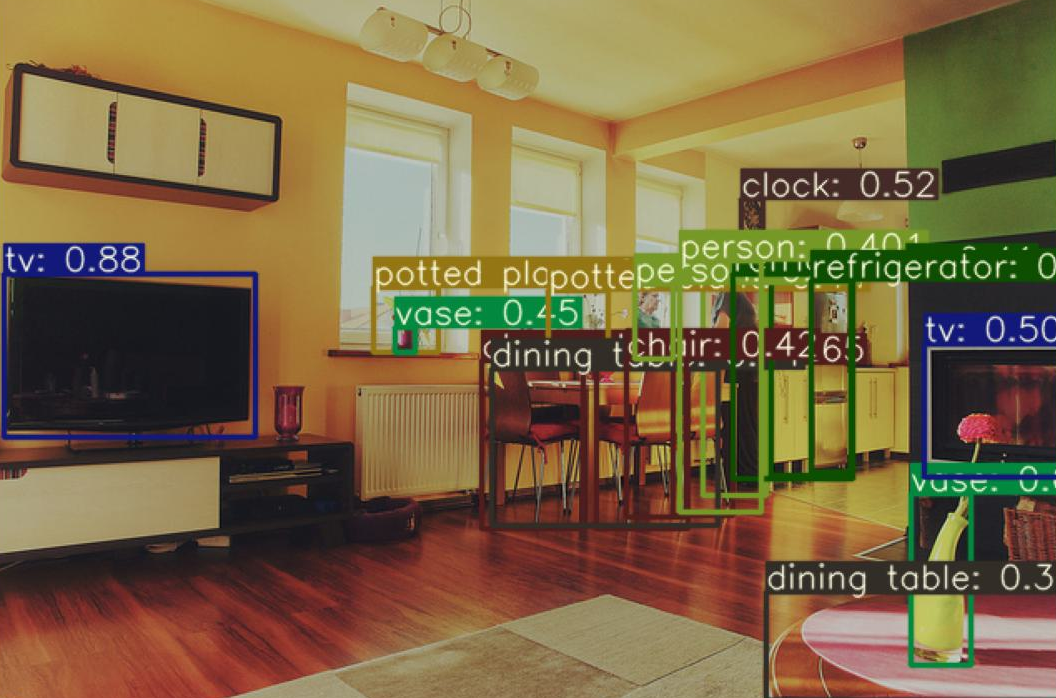}
     \end{minipage}
     \begin{minipage}{0.3\textwidth}
         \includegraphics[width=\textwidth]{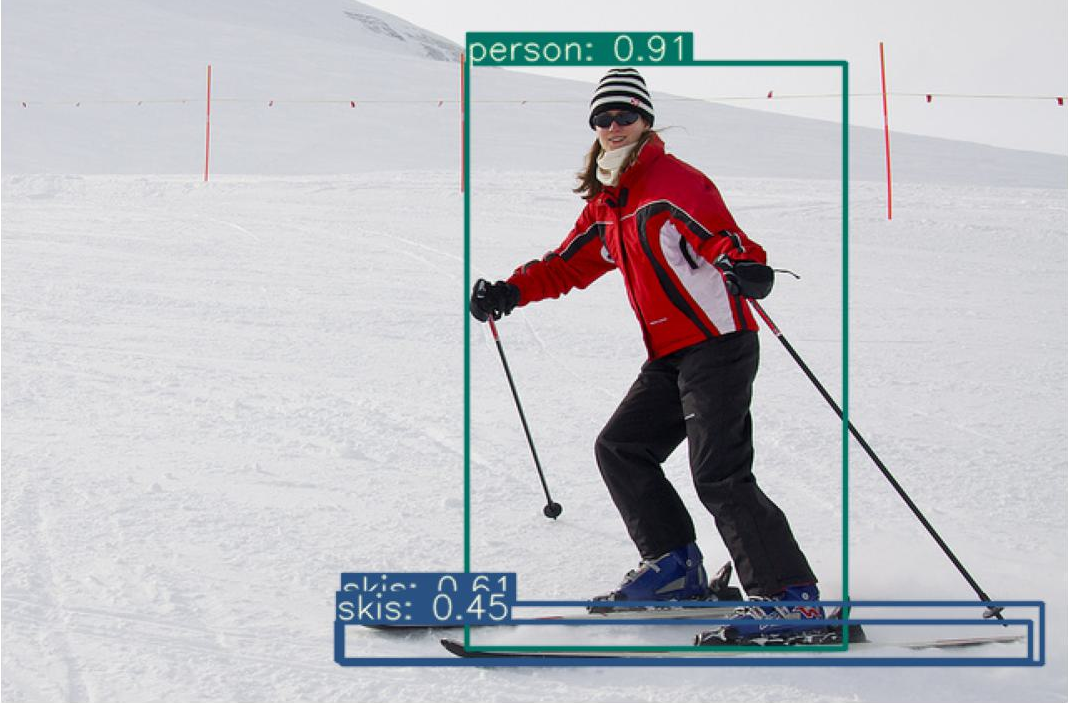}
     \end{minipage}
     
      \caption{Example of results obtained with our reproduced RTMDet on MS COCO 2017.} %
      \label{fig:RTMDet}
\end{figure*}

Our implementation gave poor results for ViTDet and DETR.
Section \ref{sec:discussion:reproductibility} goes into further detail about the different problems we faced with these models.
Our implementation of RTMDet achieved comparable results to its original paper, as shown in table \ref{tab:result_comp_originaux}.
While the latency may initially appear to be comparatively low, it should be noted that the FPS reported by \cite{RTMDet} does not include post-processing latency.
This setup, while facilitating fair comparisons between models with similar post-processing steps, artificially enhances performances of anchor-free models as they are typically used alongside post-processing in practice.
Additionally, this unfairly biases against methods such as transformer-based detection, lacking such post-processing steps.
Figure \ref{fig:RTMDet} shows examples of qualitative results obtained by our implementation of RTMDet on the validation set of MS COCO 2017.
\newline
\newline
Results obtained for our implementation of YOLOv7 are presented in table \ref{tab:result_comp_originaux_yolo}. 
Accuracy performances presented in the original article could not be matched by our own implementation.
This difference can realistically be associated to the \textit{bag-of-freebies} used in the original article, as this work didn't implement any feature within it.
On the other hand, speed performances declared in the original paper could be reproduced, albeit on different hardware, and even exceeded at greater batch size.
It is important to note that, as detailed in \ref{sec:discussion:reproductibility}, our implementation of YOLOv7 uses YOLOv7's original testing and training pipeline.
Therefore, performances obtained with this implementation should not be compared directly to other results presented in this paper.

\begin{table}[ht]
    \centering
    \caption{Model performances (original vs our reproduction) for RTMDet. The values for the original implementation of RTMDet are taken from the original article and do not include time taken by the post-processing steps.}
    \begin{tabular}{lllr}
    \toprule
    \textbf{Model} & \textbf{GPU} & \textbf{mAP} & \textbf{FPS} \\
    \midrule
    RTMDet~\cite{RTMDet} & RTX-3090 (FP16) & 44.6\% & \mbox{*}161.0 \\
    RTMDet (ours) & TitanX & 44.4\% & 44.0 \\
    RTMDet (ours) & Quadro & 44.4\% & 59.6 \\
    RTMDet (ours) & A6000 & 44.4\% & 63.1 \\
    RTMDet (ours) & RTX-3080Ti & 44.4\% & 129.9 \\
    RTMDet (ours) & RTX-4090 & 44.4\% & 160.8 \\
    \bottomrule
    \end{tabular}
    \label{tab:result_comp_originaux}
\end{table}

\begin{table}[ht]
    \centering
    \caption{Model performances (original vs our reproduction) for YOLOv7. The values for the original implementation of YOLOv7 are taken from the original article.}
    \begin{tabular}{lllrr}
    \toprule
    \textbf{Model} & \textbf{GPU} & \textbf{mAP} & \textbf{FPS} & \textbf{Batch Size} \\
    \midrule
    YOLOv7~\cite{YOLOv7} & V100 & 51.4\% & \mbox{*}161 & 1 \\
    YOLOv7 (ours) & RTX-4090 & 44.6\% & 232 & 1 \\
    YOLOv7 (ours) & RTX-4090 & 44.6\% & 27 & 16 \\
    YOLOv7 (ours) & RTX-4090 & 44.6\% & 14 & 32 \\
    \bottomrule
    \end{tabular}
    \label{tab:result_comp_originaux_yolo}
\end{table}

\begin{figure}[ht]
\includegraphics[width=8cm]{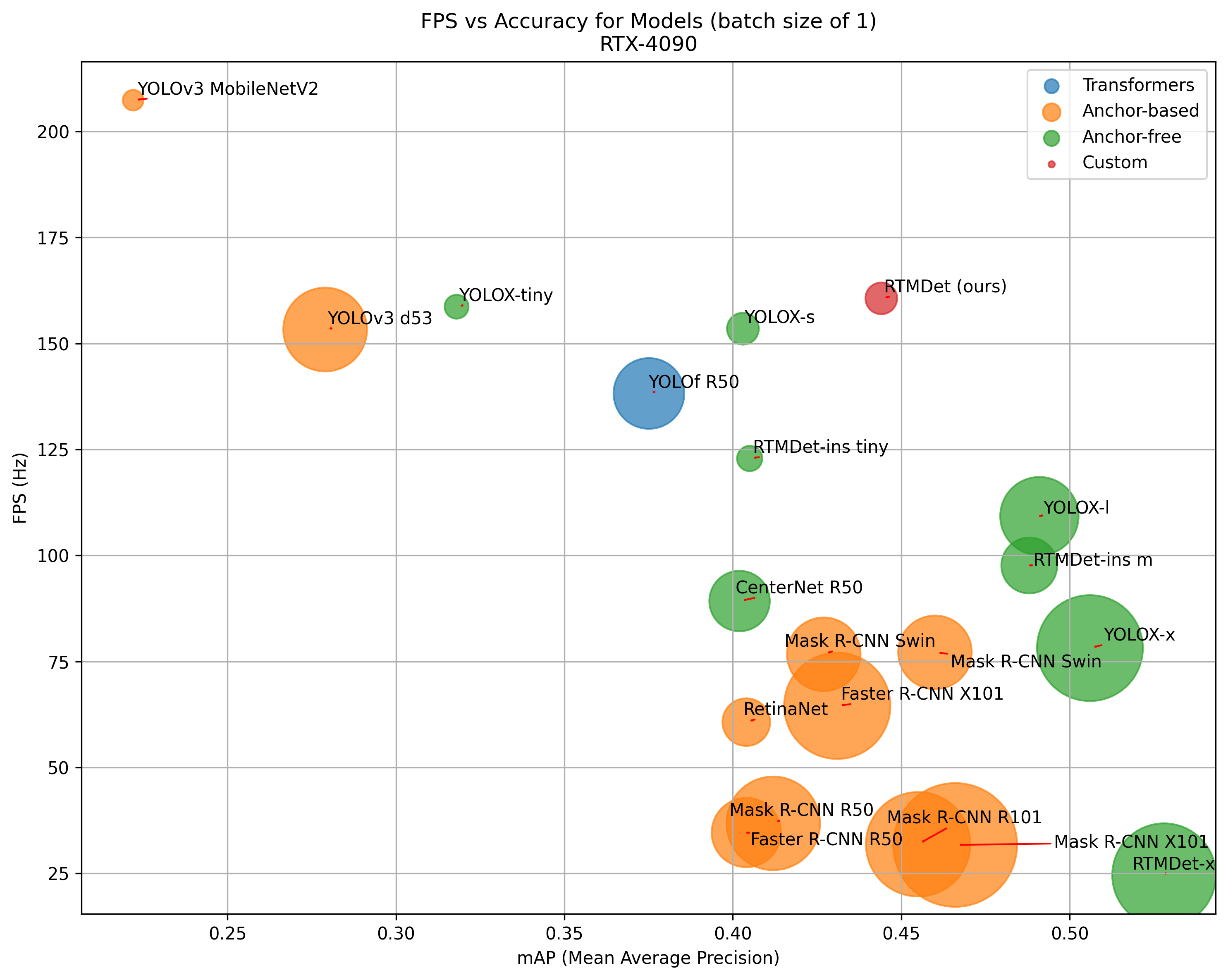}
\caption{Accuracy and frame rate of different models on RTX-4090 with a batch size of 1.}
\label{fig:batch_1}
\end{figure}

\begin{figure}[ht]
\includegraphics[width=8cm]{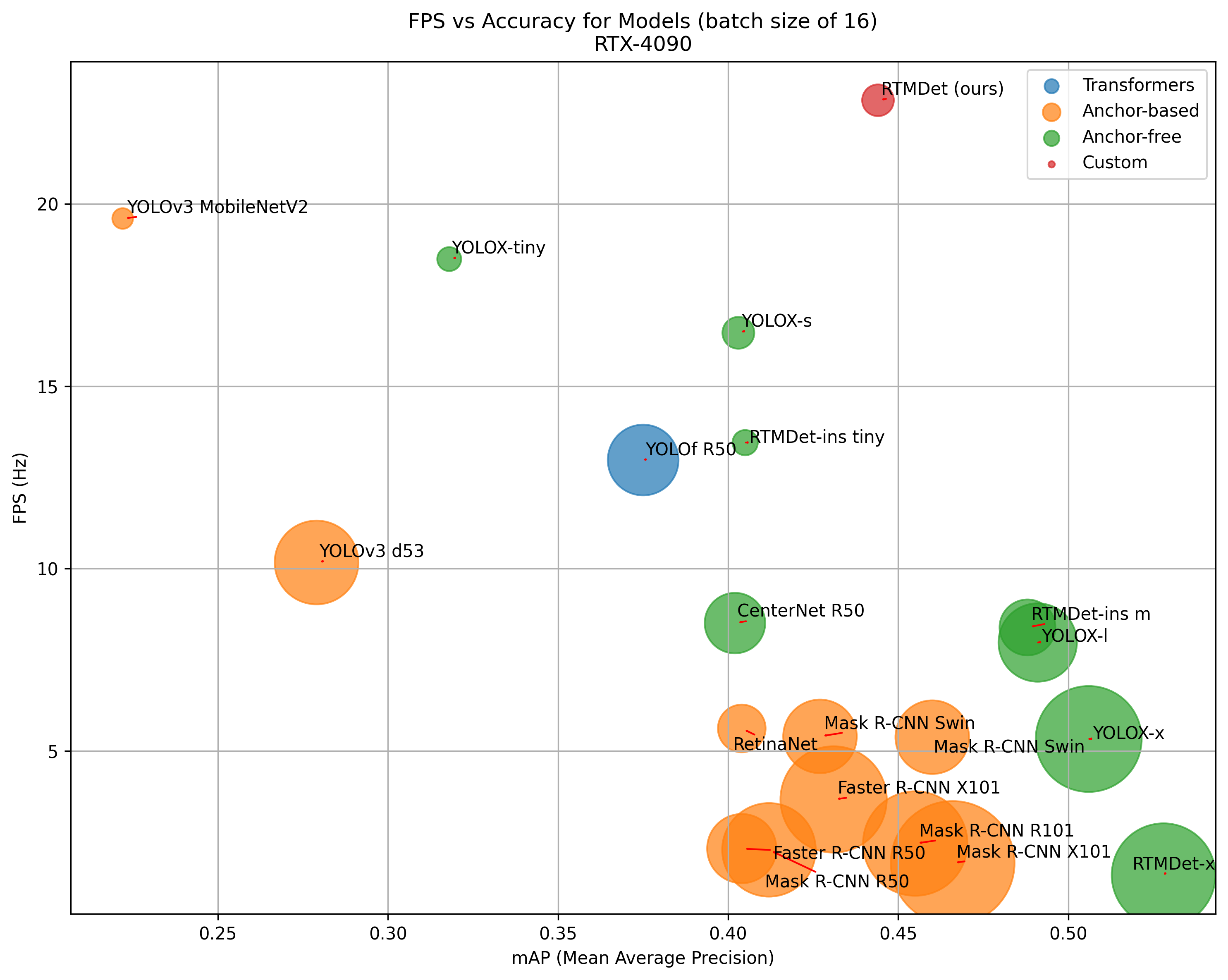}
\caption{Accuracy and frame rate of different models on RTX-4090 with a batch size of 16.}
\label{fig:batch_16}
\end{figure}

\begin{figure}[ht]
\includegraphics[width=8cm]{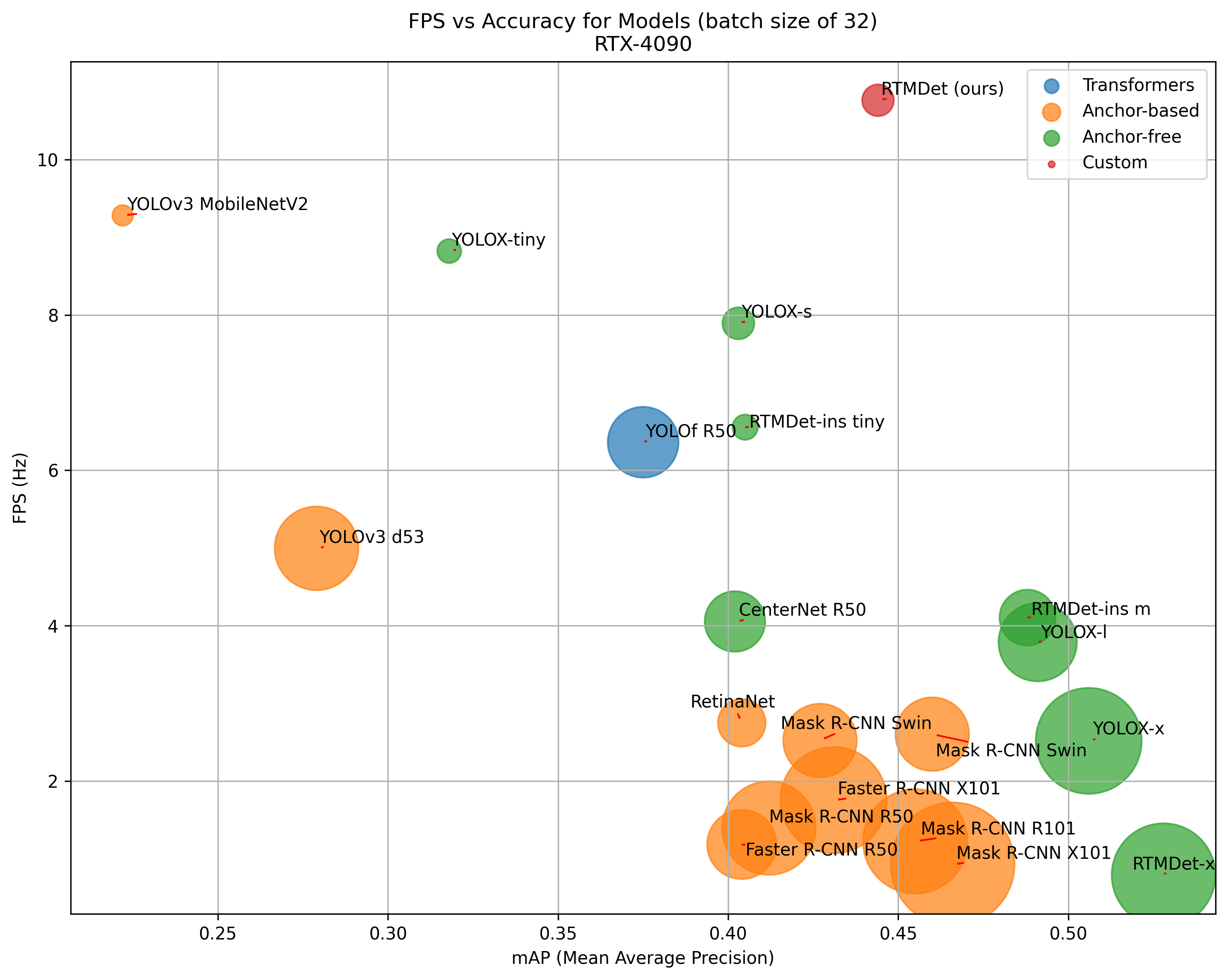}
\caption{Accuracy and frame rate of different models on RTX-4090 with a batch size of 32.}
\label{fig:batch_32}
\end{figure}

Despite not achieving competitive results with our implementations of ViTDet and DETR, we nonetheless report inference times for both models in table \ref{tab:vitdet_fps} and in table \ref{tab:detr_fps} respectively.
\\
\\
ViTDet, after 47 epochs out of the 100 planned, achieved some weak performances, reported in table \ref{tab:result_vitdet}.
Although the training process is still ongoing, those partial results still offer valuable insights.
Notably, we notice a gradation of the performances with respect to object size.
Large objects achieve an average precision of nearly 20\%, while medium objects reach around 7\%, and small objects only about 1\%.
This disparity shows the challenge of learning smaller objects compared to larger ones.
We further discuss these results in section \ref{sec:discussion:reproductibility}.
\\
\\
The implementation of DETR did not converge and has a mAP of 0\%.
The rest of this section will focus on the results obtained with the produced testing pipeline on models mentioned in section \ref{sec:method:models} as we feel it is our greatest contribution.
\\

\begin{table}[htpb]
    \centering
    \caption{Frame per seconds across considered GPU of our ViTDet implementation for a batch size of 32.}
    \begin{tabular}{lr}
        \toprule
        \textbf{GPU} & \textbf{FPS} \\
        \midrule
        TitanX & 0.2 \\
        Quadro & 0.7 \\
        A6000 & 1.0 \\
        RTX-3080Ti & 1.1\\
        RTX-4090 & 2.0\\
        \bottomrule
    \end{tabular}
    \label{tab:vitdet_fps}
\end{table}

\begin{table}[ht]
    \centering
    \caption{Model performances of our implementation of ViTDet after 47 (out of 100) epochs of training.}
    \begin{tabular}{lllr}
    \toprule
    \textbf{Metric} & \textbf{Value} \\
    \midrule
    mAP  & 0.095 \\
    AP50 & 0.186 \\
    AP75 & 0.090 \\
    APs  & 0.010 \\
    APm  & 0.065 \\
    APl  & 0.200 \\
    \bottomrule
    \end{tabular}
    \label{tab:result_vitdet}
\end{table}

\begin{table}[htbp]
    \centering
    \caption{Frame per seconds achieved by our implementation of DETR on a RTX-4090 across multiple batch size.}
    \begin{tabular}{lr}
        \toprule
        \textbf{Batch size} & \textbf{FPS} \\
        \midrule
        1 & 115.4 \\
        16 & 11.7 \\
        32 & 5.8 \\
        \bottomrule
    \end{tabular}
    \label{tab:detr_fps}
\end{table}

\subsection*{MMDet pretrained models}

We report the performance metrics for pretrained models from MMDetection in appendix \ref{sec:annex:respretrained}.
Table \ref{tab:Time_1_MMDet} shows frame-per-second performances of selected pretrained models.
Various attention-based models' inference time (DETR and DINO families) could not be estimated properly, those models require more specific data that our pipeline does not yet fully support.
Such models do offer competitive accuracy performances, but, since it is currently difficult for non-industrial users to access hardware similar to the high-end GPUS tested (RTX-4090), common usage of DINO and DETR models (as implemented in MMDetection) for real-time object detection seems impractical because of the memory requirements for transformer based models.
For transformer-based models, YOLOf R50 exhibits comparable speed performances to anchor-free and anchor-based architectures.
\newline
\newline
Table \ref{tab:Time_1_MMDet} shows improved speed performances for all models when computing resources increase, which is expected.
We also provide the achieved mAP on all selected models in table \ref{tab:mAP_Comp_MMDet}.
\newline
\newline
Additionally, figures \ref{fig:batch_1}, \ref{fig:batch_16} and \ref{fig:batch_32}\footnote{The exact numbers used to generate these graphs are available in appendix \ref{sec:annex:respretrained}} reveals that increasing the batch size reduces speed performances for every model studied.
More importantly, not all models seem to be affected in the same way by an increased batch size.
Indeed, YOLOx and RTMDet family models maintain better speed performances with an increased batch size than other models tested.
Broadly speaking, figures \ref{fig:batch_1}, \ref{fig:batch_16} and \ref{fig:batch_32} imply that lighter models are less affected by a larger batch size (large models agglomerate to the bottom of the graphs).
\newline
\newline
On the flip side, the larger models depicted in the same figure tend to showcase higher levels of accuracy.
In essence, the figure illustrates a relative trade-off between accuracy and inference time.
Models belonging to the YOLOx and RTMDet families appear to strike a better balance between accuracy and speed compared to other models tested, at least within the range of GPUs examined.
This observation extends to our RTMDet implementation, where an aggressive pre-filtering strategy results in lighter and more efficient NMS and thresholding procedures.
This approach appears to lead to better-amortized latency, as demonstrated in Figures \ref{fig:batch_16} and \ref{fig:batch_32}, emphasizing the crucial role of post-processing in latency measurement.

\section{Discussion}

\subsection*{Reproducibility}
\label{sec:discussion:reproductibility}
The following considers in more detail the issues linked to the reproducibility of selected benchmark models.

\subsection*{Yolov7}

Yolov7 truly shines for out-of-the-box usage.
Indeed, code, weights and usage instructions are easily accessible and user-friendly --- including both Docker setup instructions and Google Colab premade setup.
These instructions also include the necessary explanation to use features within the \textit{bag-of-freebies} with Yolov7 premade models. 
\newline
\newline
Yolov7, taken as a product, is an excellent model.
However, the same cannot be said for developers: details given within the article and the code repository are insufficient for reproducibility or model customization.
First, the article is highly dependent on prior knowledge of previous YOLO-family models.
For example, it is mentioned that the architecture is based on YOLOv5's and includes YOLOR's implicit knowledge, but no details are given on how this is implemented (whether in the article or the code repository).
Searching previous YOLO papers and architectures (mainly YOLO3~\cite{YOLOv3}, YOLOR~\cite{YOLOR} and YOLOv5~\cite{YOLOv5}) is essential to understand and reproduce YOLOv7 from scratch.
Other articles are referenced as the source of some code blocks in the code itself (for example, YOLOv7 uses a SPPCSPN\footnote{Cross Stage Partial Networks Spatial pyramid pooling layer from \url{https://github.com/WongKinYiu/CrossStagePartialNetworks}, consulted April 21st 2024.} layer) but not mentioned within the article or the usage instructions.
Additionally, the article neglects important information necessary for the reproduction of training conditions, including the data augmentation and the loss function used.
\newline
\newline
Moreover, YOLOv7 sometimes neglects to use conventional PyTorch or TensorFlow code structure in favor of made-from-scratch implementations of corresponding functionalities.
For example, YOLOv7 uses a custom data loader for its training, a custom data augmentation pipeline and a custom learning rate scheduler.
This wouldn't be a problem by itself, but the fact is that YOLOv7's training environment is highly dependent on said custom structure.
In that way, it is quite challenging to reuse the same training regiment as YOLOv7 with a custom reimplementation of the model.
In the same way, YOLOv7's pre-trained weights are defined with a custom pickle\footnote{\url{https://docs.python.org/3/library/pickle.html}, consulted April 26th 2024.} format instead of \texttt{pth} as commonly used in PyTorch, complexifying its usage outside of YOLOv7's premade architecture, as it depends on having specific Python modules at specific paths.
The complexity of reproduction is further exacerbated by the absence of details in the original article for any such custom feature.
\newline
\newline
Finally, the included \textit{bag-of-freebies} is interwoven within the model's code and acts as a lot of background noise when trying to understand the model itself.
Which \textit{bag-of-freebies}' functionalities and with what parameters were used to obtain the results presented in the original paper is also mostly left unanswered.
In the end, YOLOv7's training took 8 days.

\subsection*{RTMDet}

The MMdetection toolbox offers an elegant, easily extendable modular design.
This characteristic makes it a fitting choice to reimplement not only RTMDet, but most of our selected algorithms.
Working within the toolbox conventions enabled us to leverage its rich ecosystem of data processing, data augmentation, loss function and post-processing utilities.
Especially for RTMDet, as every tool mentioned in the paper was available directly within the repository, which allowed us to focus on the reimplementation of the model itself.
\newline
\newline
While MMdetection toolbox's excellent software engineering practices facilitate easy extensibility, they introduce a significant level of abstraction, making it challenging to locate actual implementations.
This, combined with some lack of details in the paper, made some parts of the reimplementation more difficult.
Notably, the CSPNeXt backbone is not entirely described in the paper.
Furthermore, discrepancies were observed between the actual loss function utilized in the repository and the well-described variant presented in the paper.
Curiously, the implementation uses equation \ref{eq:loss_rtmdet} as its loss function, which is only briefly described as the baseline in the paper.
Training took 7 days and happened without incident.

\subsection*{ViTDet}
Some important hyperparameters necessary to reproduce ViTDet were absent in the original paper.
First, the paper mentions the usage of a learning rate step during training without further details.
Searching the official code repository shows a multiplicative step of 0.1 for iterations 163889 to 177546.
Additionally, the original paper does describe the SFP architecture, but omits to mention that the intermediary embedding between convolution and deconvolution layers reduces channel width by half.
The article does mention a ViT backbone based on Masked Autoencoder~\cite{Masked}, but modifications are made to it in the official code repository for ViTDet which are only briefly mentioned by the original authors.
Besides these oversights, reproducibility was mostly uneventful.
\newline
\newline
The training itself was more time-consuming than expected.
At the time of writing this paper, it is still not finished and goes at a rate of 14 iterations per day (on a A6000 GPU).
At this rate, the total training time will be more than 7 days. 
Due to time constraints and occasional divergences in training, it became impossible to complete the entire training schedule planned in the article.
Nonetheless, we report the inference time and accuracy of our implementation of ViTDet in tables \ref{tab:vitdet_fps} and \ref{tab:result_vitdet}.
The results shown are obtained after 47 epochs.
Our implementation achieved a mAP of approximately 10\% by that point, confirming our network is learning, albeit at a prohibitively slow rate for our time constraints.

\subsection*{DETR}

DETR is conceptually simple and paired with a well-detailed article.
Reproduction using exclusively the original paper is achievable.
Notably, the batch size had to be adjusted to 40 instead of 64 due to memory limitations.
Otherwise, hyperparameters (see table \ref{tab:hyperparameters}) presented in the article were used directly.
However, issues arose at training.
After the intended 300 epochs (6 days of training time), the network always predicted the \textit{No object} label.
Looking at the official code, a few omissions within the paper were found.
First, the normalization layer within the transformers is in reality placed before the attention mechanism instead of after.
The positional encoding added to the image is only applied to the non-padded section of the input.
We tried implementing these changes, but the network still consistently predicted the \textit{No object} label.
In fact, the learning rate schedule used in the original article seem too simple to achieve appreciable results.
Pairing this with the lack of data augmentation makes it seem like the training process presented is lacking refinement, which hinders the speed at which the network learns.
The model still learns after the 300 epochs originally used, so we are hopeful that we could achieve similar performances to the original paper, but with significantly higher training costs.
Nonetheless, we report the inference speed of our implementation in table \ref{tab:detr_fps}, and found it competitive in terms of inference time.

\section{Conclusion}

In this work, four state-of-the-art models (YOLOv7, RTMDet, ViTDet and DETR) were selected for reproduction to determine if such models offer sufficient accuracy and speed in a non-industrial environment.
Overall, only RTMDet and YOLOv7 could offer significant performances.
The two other models were found to be too computationally expansive for our hardware, much more than what could be expected from the original papers.
Moreover, papers for YOLOv7, ViTDet and DETR were found to be insufficient for reproducibility purposes, lacking fundamental information on both the architecture or training context.
\newline
\newline
A unified training and testing pipeline built on MMDetection was produced and used to evaluate 30 different pre-trained models offered by MMOpenLab.
Experimentation showed the trade-offs between model accuracy, speed and size.
For the context of real-time object detection, models based on an anchor-free architecture (mostly RTMDet and YOLOx) achieve a better compromise between accuracy, speed and size than tested anchor-based or attention-based models.
\newline
\newline
Future works could include testing with additional hardware (including Nvidia Jetson and Orin GPUs to investigate the use of these models in embedded settings) using the same unified pipeline.
Additional hardware could help to support speed performance estimation for the more demanding architectures (notably DETR and DINO).
We also wish to explore the impact of mixed precision on model inference speed.
Moreover, MMDetection offers various other models besides those considered in this work, which were initially excluded due to time constraints.
Expanding our benchmark to include a wider range of models beyond those available in MMDetection, we aim to incorporate models provided by Hugging Face as well.\footnote{\url{https://huggingface.co/}, consulted on April 28th 2024}
The built pipeline is found to be flexible and reliable; model benchmarking on a larger scale based on this infrastructure seems to be an achievable step forward for this work.
We also aim to test additional models beyond just object detection, such as panoptics, 3D object detection and human pose estimation, with our benchmarking pipeline.

\section*{Acknowledgement}

We would also like to acknowledge the generous support of NVIDIA Corporation for providing a Quadro RTX 8000 GPU, which enhanced the computational resources used in this research.
Thanks to the Northern Robotics Laboratory (norlab)\footnote{{https://norlab.ulaval.ca/}, consulted April 21st 2024} from Laval University for access to computing resources. Thanks to David-Alexandre Duclos for allowing us to use his TitanX GPU.

\printbibliography
\onecolumn
\appendix
\section*{Appendix}

\section{Hyperparameters used}

\begin{table*}[h]
    \centering
    \caption{List of used hyperparameters for implemented models.}
    \begin{tabular}{lllll}
    \toprule
    \textbf{Hyperparameter} & \textbf{YOLOv7} & \textbf{RTMDet} & \textbf{ViTDet} & \textbf{DETR} \\
    \midrule
    Optimizer & SGD & AdamW & AdamW & AdamW \\
    Base Learning Rate & 0.01 & 0.00125 & 0.0001 & \makecell{0.00001 for backbone \\ 0.0001 for Transformer} \\
    Learning Rate Schedule & \makecell{Sinusoidal ramp \\ from 0.01 to 0.1} & \makecell{Flat (150 epochs) \\ then Cosine} & \makecell{Multiplicative step \\ of 0.1 for iterations \\ 163889 to 177546} & \makecell{Multiplicative step of 0.1 \\ after 200 epochs} \\
    Weight Decay & 0.0005 & \makecell{0.05 (0 for bias \\ and normalization} & 0.01 & 0.0001 \\
    Optimizer Momentum & 0.937 & 0.9 & - & - \\
    Batch size & 64 & 80 & 24 (per GPU) & 40 \\
    Dropout Rate & - & - & 0.1 & 0.1 \\
    Maximum iterations & - & - & 184375 & - \\
    Training Epochs & 300 & 300 & 100 & 300 \\
    Warmup Iterations & 3 & 1000 & 250 & - \\
    Warmup momentum & 0.8 & - & 0.001 & - \\
    EMA Decay & - & 0.9998 & - & - \\
    Input Size & 640 $\times$ 640 & 640 $\times$ 640 & 1024 $\times$ 1024 & - \\
    \hline
    Loss & \makecell{BCELoss \\ 5\% box gain \\ 30\% cls gain \\ 70\% obj gain} & \makecell{QFL and GIoU} & \makecell{Classification: \\ Cross-Entropy \\ Bounding boxes: \\Smooth1Loss} & \makecell{$l_1$ loss with weight 5 \\ GIoU loss with weight 2}\\
    \hline
    Augmentation & \makecell{Mosaic (100\%) \\ MixUp (15\%) \\ Copy paste (15\%) \\ Translation (20\%) \\ Scale (90\%) \\ Flip Left-Right (50\%) \\ HSV: \\ 1.5\% hue fraction \\ 70\% saturation fraction \\ 40\% value fraction} & \makecell{Mosaic and MixUp \\ (first 280 epochs) \\ LSJ(~\cite{DETECTRON2}, ~\cite{LAUG}) \\ (last 20 epochs)} & \makecell{Mosaic and MixUp \\ LSJ (~\cite{DETECTRON2}, ~\cite{LAUG})} & Crop and scale \\
    \bottomrule
    \end{tabular}
    \label{tab:hyperparameters}
\end{table*}

\newpage
\section{MMDetection pretrained model performances}

\begin{table*}[ht]
\centering
\caption{Comparative frame per second performances (batch size of 1)}
\begin{tabular}{lrrrrrr}
\toprule
Model & Size (MiB) & TitanX & Quadro & A6000 & RTX-3080 & RTX-4090 \\
\midrule
Mask R-CNN R101 & 367.82 & 6.60 & 10.73 & 14.62 & 20.39 & 31.46 \\
Mask R-CNN R50 & 295.17 & 7.89 & 12.69 & 19.69 & 21.25 & 37.70 \\
Mask R-CNN X101 & 516.61 & 4.35 & 8.51 & 14.08 & 15.21 & 32.23 \\
R-CNN Faster R50 & 161.72 & 9.31 & 13.26 & 14.66 & 26.11 & 34.01 \\
Centernet R50 & 123.39 & 24.68 & 32.78 & 30.30 & 66.25 & 91.61 \\
Cond. DETR R50 & 166.02 & - & - & - & - & - \\
DAB DETR R50 & 166.99 & - & - & - & - & - \\
Deform. DETR R50 & 153.24 & - & - & - & - & - \\
D. DETR R50 (2 stages) & 157.42 & - & - & - & - & - \\
DETR R50 & 158.80 & - & - & - & - & - \\
DINO-4 R50 & 182.17 & - & - & - & - & - \\
DINO-5 Swin & 836.67 & - & - & - & - & - \\
Faster R-CNN Swin & 380.91 & 7.18 & 14.24 & 24.77 & 27.51 & 61.00 \\
Mask R-CNN Swin & 182.54 & 15.80 & 25.11 & 30.21 & 39.04 & 73.12 \\
Mask R-CNN Swin (Crop) & 182.54 & 14.83 & 29.18 & 35.52 & 35.98 & 75.01 \\
Retinanet & 76.62 & 16.56 & 22.37 & 22.34 & 41.36 & 60.50 \\
RTMDet-x & 362.30 & 7.49 & 10.74 & 10.89 & 15.94 & 24.49 \\
RTMDet-ins m & 105.38 & 26.30 & 36.62 & 33.68 & 69.06 & 91.82 \\
RTMDet-ins tiny & 21.58 & 31.44 & 44.84 & 41.68 & 92.19 & 115.60 \\
YOLOf R50 & 168.71 & 37.91 & 50.48 & 48.71 & 99.88 & 136.11 \\
YOLOv3 d53 & 236.52 & 31.83 & 44.43 & 66.41 & 73.92 & 151.82 \\
YOLOv3 MobileNetV2 & 14.41 & 50.39 & 69.21 & 66.20 & 154.00 & 208.64 \\
YOLOX-l & 207.06 & 28.98 & 39.33 & 36.36 & 59.36 & 107.36 \\
YOLOX-s & 34.30 & 41.49 & 54.37 & 54.39 & 110.49 & 142.14 \\
YOLOX-tiny & 19.35 & 40.60 & 54.47 & 53.27 & 111.78 & 149.90 \\
YOLOX-x & 378.31 & 18.17 & 24.09 & 38.23 & 41.08 & 76.39 \\
\bottomrule
\end{tabular}
\label{tab:Time_1_MMDet}
\end{table*}

\begin{table*}[ht]
\centering
\caption{Comparative frame per second performances (batch size of 16).}
\begin{tabular}{lrrrrrr}
\toprule
Model & Size (MiB) & TitanX & Quadro & A6000 & RTX-3080 & RTX-4090 \\
\midrule
Mask R-CNN R101 & 367.82 & 0.36 & 0.88 & 1.55 & 1.45 & 2.45 \\
Mask R-CNN R50 & 295.17 & 0.43 & 1.03 & 1.47 & 1.39 & 2.70 \\
Mask R-CNN X101 & 516.61 & 0.25 & 0.67 & 1.23 & 1.16 & 2.09 \\
R-CNN Faster R50 & 161.72 & 0.65 & 0.97 & 1.11 & 1.91 & 2.51 \\
Centernet R50 & 123.39 & 2.62 & 3.73 & 5.15 & 5.25 & 8.92 \\
Cond. DETR R50 & 166.02 & - & - & - & - & - \\
DAB DETR R50 & 166.99 & - & - & - & - & - \\
Deform. DETR R50 & 153.24 & - & - & - & - & - \\
D. DETR R50 (2 stages) & 157.42 & - & - & - & - & - \\
DETR R50 & 158.80 & - & - & - & - & - \\
DINO-4 R50 & 182.17 & - & - & - & - & - \\
DINO-5 Swin & 836.67 & - & - & - & - & - \\
Faster R-CNN Swin & 380.91 & 0.42 & 1.17 & 1.92 & 2.22 & 3.25 \\
Mask R-CNN Swin & 182.54 & 0.82 & 2.14 & 2.84 & 2.92 & 5.28 \\
Mask R-CNN Swin (Crop) & 182.54 & 0.94 & 2.21 & 2.89 & 2.56 & 5.25 \\
Retinanet & 76.62 & 1.48 & 2.53 & 3.41 & 3.58 & 5.61 \\
RTMDet-x & 362.30 & 0.53 & 0.72 & 0.83 & 1.02 & 1.55 \\
RTMDet-ins m & 105.38 & 2.25 & 3.62 & 4.43 & 5.32 & 8.48 \\
RTMDet-ins tiny & 21.58 & 3.89 & 6.25 & 6.34 & 9.16 & 13.62 \\
YOLOf R50 & 168.71 & 3.92 & 5.45 & 7.54 & 8.75 & 12.49 \\
YOLOv3 d53 & 236.52 & 2.86 & 4.13 & 6.08 & 6.01 & 9.73 \\
YOLOv3 MobileNetV2 & 14.41 & 5.03 & 8.52 & 10.17 & 12.72 & 19.22 \\
YOLOX-l & 207.06 & 2.31 & 3.25 & 4.51 & 4.62 & 7.75 \\
YOLOX-s & 34.30 & 5.41 & 7.26 & 7.83 & 9.51 & 15.61 \\
YOLOX-tiny & 19.35 & 6.35 & 8.29 & 8.54 & 10.56 & 17.99 \\
YOLOX-x & 378.31 & 1.55 & 2.03 & 3.29 & 2.94 & 5.22 \\
\bottomrule
\end{tabular}
\label{tab:Time_16_MMDet}
\end{table*}

\begin{table*}[ht]
\centering
\caption{Comparative frame per second performances (batch size of 32).}
\begin{tabular}{lrrrrrr}
\toprule
Model & Size (MiB) & TitanX & Quadro & A6000 & RTX-3080 & RTX-4090 \\
\midrule
Mask R-CNN R101 & 367.82 & 0.42 & 0.75 & - & - & 1.10 \\
Mask R-CNN R50 & 295.17 & 0.52 & 0.89 & - & - & 1.42 \\
Mask R-CNN X101 & 516.61 & 0.32 & 0.54 & - & - & 0.98 \\
R-CNN Faster R50 & 161.72 & 0.50 & 0.61 & - & - & 1.22 \\
Centernet R50 & 123.39 & 1.27 & 1.81 & 2.47 & 2.82 & 4.15 \\
Cond. DETR R50 & 166.02 & - & - & - & - & - \\
DAB DETR R50 & 166.99 & - & - & - & - & - \\
Deform. DETR R50 & 153.24 & - & - & - & - & - \\
D. DETR R50 (2 stages) & 157.42 & - & - & - & - & - \\
DETR R50 & 158.80 & - & - & - & - & - \\
DINO-4 R50 & 182.17 & - & - & - & - & - \\
DINO-5 Swin & 836.67 & - & - & - & - & - \\
Faster R-CNN Swin & 380.91 & 0.2 & 0.63 & 1.09 & 1.01 & 1.73 \\
Mask R-CNN Swin & 182.54 & 0.43 & 1.12 & 1.43 & 1.50 & 2.48 \\
Mask R-CNN Swin (Crop) & 182.54 & 0.44 & 1.09 & 1.33 & 1.45 & 2.47 \\
Retinanet & 76.62 & 0.75 & 1.27 & 1.74 & 1.70 & 2.73 \\
RTMDet-x & 362.30 & 0.26 & 0.36 & 0.43 & 0.50 & 0.76 \\
RTMDet-ins m & 105.38 & 1.11 & 1.84 & 2.44 & 2.63 & 4.00 \\
RTMDet-ins tiny & 21.58 & 1.94 & 3.12 & 3.40 & 4.67 & 6.53 \\
YOLOf R50 & 168.71 & 1.98 & 2.78 & 4.09 & 4.49 & 6.18 \\
YOLOv3 d53 & 236.52 & 1.44 & 2.08 & 3.17 & 2.98 & 4.87 \\
YOLOv3 MobileNetV2 & 14.41 & 2.52 & 4.25 & 5.11 & 6.39 & 9.01 \\
YOLOX-l & 207.06 & 1.17 & 1.65 & 2.35 & 2.39 & 3.77 \\
YOLOX-s & 34.30 & 2.70 & 3.64 & 4.08 & 4.82 & 7.60 \\
YOLOX-tiny & 19.35 & 3.18 & 4.13 & 4.36 & 5.14 & 8.66 \\
YOLOX-x & 378.31 & 0.78 & 1.08 & 1.68 & 1.47 & 2.47 \\
\bottomrule
\end{tabular}
\label{tab:Time_32_MMDet}
\end{table*}

\label{sec:annex:respretrained}

\begin{table*}[ht]
\centering
\caption{Accuracy performances for MMDetection pretrained networks.}
\begin{tabular}{lrrrrrrr}
\toprule
Model & Size (MiB) & mAP & AP50 & AP75 & APs & APm & APl \\
\midrule
Mask R-CNN R101 & 367.82 & 0.46 & 0.64 & 0.50 & 0.28 & 0.49 & 0.59 \\
Mask R-CNN R50 & 295.17 & 0.41 & 0.59 & 0.45 & 0.24 & 0.44 & 0.54 \\
Mask R-CNN X101 & 516.61 & 0.47 & 0.65 & 0.51 & 0.29 & 0.50 & 0.60 \\
R-CNN Faster R50 & 161.72 & 0.40 & 0.59 & 0.44 & 0.23 & 0.44 & 0.53 \\
Centernet R50 & 123.39 & 0.40 & 0.58 & 0.44 & 0.23 & 0.45 & 0.52 \\
Cond. DETR R50 & 166.02 & 0.41 & 0.62 & 0.44 & 0.20 & 0.44 & 0.60 \\
DAB DETR R50 & 166.99 & 0.42 & 0.63 & 0.45 & 0.22 & 0.46 & 0.61 \\
Deform. DETR R50 & 153.24 & 0.44 & 0.63 & 0.49 & 0.27 & 0.48 & 0.59 \\
Deform. DETR R50 (2 stages) & 157.42 & 0.47 & 0.66 & 0.51 & 0.30 & 0.50 & 0.62 \\
DETR R50 & 158.80 & 0.40 & 0.60 & 0.42 & 0.18 & 0.44 & 0.59 \\
DINO-4 R50 & 182.17 & 0.50 & 0.68 & 0.55 & 0.33 & 0.53 & 0.64 \\
DINO-5 Swin & 836.67 & 0.58 & 0.77 & 0.64 & 0.42 & 0.62 & 0.74 \\
Faster R-CNN Swin & 380.91 & 0.43 & 0.64 & 0.47 & 0.26 & 0.47 & 0.56 \\
Mask R-CNN Swin & 182.54 & 0.43 & 0.65 & 0.47 & 0.26 & 0.46 & 0.57 \\
Mask R-CNN Swin (Crop) & 182.54 & 0.46 & 0.68 & 0.50 & 0.30 & 0.49 & 0.60 \\
Retinanet & 76.62 & 0.40 & 0.60 & 0.43 & 0.22 & 0.45 & 0.56 \\
RTMDet-x & 362.30 & 0.53 & 0.70 & 0.58 & 0.36 & 0.57 & 0.69 \\
RTMDet-ins m & 105.38 & 0.49 & 0.67 & 0.53 & 0.30 & 0.54 & 0.65 \\
RTMDet-ins tiny & 21.58 & 0.40 & 0.58 & 0.44 & 0.21 & 0.45 & 0.58 \\
YOLOf R50 & 168.71 & 0.38 & 0.57 & 0.40 & 0.19 & 0.42 & 0.53 \\
YOLOv3 d53 & 236.52 & 0.28 & 0.49 & 0.28 & 0.10 & 0.30 & 0.44 \\
YOLOv3 MobileNetV2 & 14.41 & 0.22 & 0.42 & 0.22 & 0.06 & 0.24 & 0.36 \\
YOLOX-l & 207.06 & 0.49 & 0.67 & 0.53 & 0.32 & 0.54 & 0.64 \\
YOLOX-s & 34.30 & 0.40 & 0.59 & 0.43 & 0.24 & 0.44 & 0.53 \\
YOLOX-tiny & 19.35 & 0.32 & 0.49 & 0.34 & 0.12 & 0.35 & 0.47 \\
YOLOX-x & 378.31 & 0.51 & 0.68 & 0.55 & 0.32 & 0.56 & 0.67 \\
\bottomrule
\end{tabular}
\label{tab:mAP_Comp_MMDet}
\end{table*}

\begin{landscape}
\begin{table*}[ht]
\centering
\caption{Time performances, in seconds, for MMDet pretrained models with TitanX (batch size of 1).}
\begin{tabular}{lrrrrrr}
\toprule
Model & Size (MiB) & Time 1 mean & Time 1 std & Time 1 min & Time 1 median & Time 1 max \\
\midrule
Mask R-CNN R101 & 367.822 & 0.151 & 0.005 & 0.137 & 0.151 & 0.176 \\
Mask R-CNN R50 & 295.173 & 0.127 & 0.004 & 0.117 & 0.126 & 0.144 \\
Mask R-CNN X101 & 516.609 & 0.230 & 0.010 & 0.212 & 0.226 & 0.262 \\
R-CNN Faster R50 & 161.722 & 0.107 & 0.003 & 0.062 & 0.107 & 0.147 \\
Centernet R50 & 123.391 & 0.041 & 0.001 & 0.039 & 0.041 & 0.072 \\
Cond. DETR R50 & 166.023 & - & - & - & - & - \\
DAB DETR R50 & 166.989 & - & - & - & - & - \\
Deform. DETR R50 & 153.245 & - & - & - & - & - \\
Deform. DETR R50 (2 stages) & 157.421 & - & - & - & - & - \\
DETR R50 & 158.799 & - & - & - & - & - \\
DINO-4 R50 & 182.173 & - & - & - & - & - \\
DINO-5 Swin & 836.669 & - & - & - & - & - \\
Faster R-CNN Swin & 380.914 & 0.139 & 0.006 & 0.131 & 0.137 & 0.162 \\
Mask R-CNN Swin & 182.542 & 0.063 & 0.005 & 0.052 & 0.062 & 0.088 \\
Mask R-CNN Swin (Crop) & 182.542 & 0.067 & 0.007 & 0.053 & 0.066 & 0.094 \\
Retinanet & 76.619 & 0.060 & 0.002 & 0.058 & 0.060 & 0.087 \\
RTMDet-x & 362.298 & 0.133 & 0.004 & 0.126 & 0.132 & 0.156 \\
RTMDet-ins m & 105.383 & 0.038 & 0.001 & 0.035 & 0.038 & 0.052 \\
RTMDet-ins tiny & 21.579 & 0.032 & 0.001 & 0.029 & 0.032 & 0.047 \\
YOLOf R50 & 168.715 & 0.026 & 0.001 & 0.025 & 0.027 & 0.040 \\
YOLOv3 d53 & 236.519 & 0.031 & 0.001 & 0.029 & 0.031 & 0.039 \\
YOLOv3 MobileNetV2 & 14.408 & 0.020 & 0.001 & 0.018 & 0.020 & 0.031 \\
YOLOX-l & 207.056 & 0.035 & 0.001 & 0.033 & 0.034 & 0.049 \\
YOLOX-s & 34.300 & 0.024 & 0.001 & 0.023 & 0.024 & 0.034 \\
YOLOX-tiny & 19.353 & 0.025 & 0.001 & 0.023 & 0.025 & 0.036 \\
YOLOX-x & 378.314 & 0.055 & 0.003 & 0.051 & 0.054 & 0.063 \\
\bottomrule
\end{tabular}
\label{tab:TitanX_time1_MMDet}
\end{table*}
\end{landscape}

\begin{landscape}
\begin{table*}[ht]
\centering
\caption{Time performances, in seconds, for MMDet presets with TitanX (batch size of 16).}
\begin{tabular}{lrrrrrr}
\toprule
Model & Size (MiB) & Time 16 mean & Time 16 std & Time 16 min & Time 16 median & Time 16 max \\
\midrule
Mask R-CNN R101 & 367.822 & 2.772 & 0.070 & 2.021 & 2.773 & 2.863 \\
Mask R-CNN R50 & 295.173 & 2.320 & 0.064 & 2.121 & 2.324 & 2.485 \\
Mask R-CNN X101 & 516.609 & 3.934 & 0.038 & 3.792 & 3.939 & 4.007 \\
R-CNN Faster R50 & 161.722 & 1.543 & 0.023 & 1.471 & 1.539 & 1.609 \\
Centernet R50 & 123.391 & 0.381 & 0.008 & 0.368 & 0.381 & 0.398 \\
Cond. DETR R50 & 166.023 & - & - & - & - & - \\
DAB DETR R50 & 166.989 & - & - & - & - & - \\
Deform. DETR R50 & 153.245 & - & - & - & - & - \\
Deform. DETR R50 (2 stages) & 157.421 & - & - & - & - & - \\
DETR R50 & 158.799 & - & - & - & - & - \\
DINO-4 R50 & 182.173 & - & - & - & - & - \\
DINO-5 Swin & 836.669 & - & - & - & - & - \\
Faster R-CNN Swin & 380.914 & 2.387 & 0.096 & 2.185 & 2.422 & 2.533 \\
Mask R-CNN Swin & 182.542 & 1.216 & 0.040 & 0.991 & 1.221 & 1.301 \\
Mask R-CNN Swin (Crop) & 182.542 & 1.058 & 0.033 & 0.988 & 1.057 & 1.134 \\
Retinanet & 76.619 & 0.674 & 0.018 & 0.648 & 0.670 & 0.717 \\
RTMDet-x & 362.298 & 1.902 & 0.023 & 1.875 & 1.892 & 1.968 \\
RTMDet-ins m & 105.383 & 0.444 & 0.011 & 0.429 & 0.441 & 0.470 \\
RTMDet-ins tiny & 21.579 & 0.257 & 0.003 & 0.252 & 0.256 & 0.275 \\
YOLOf R50 & 168.715 & 0.255 & 0.006 & 0.249 & 0.252 & 0.279 \\
YOLOv3 d53 & 236.519 & 0.350 & 0.008 & 0.339 & 0.347 & 0.368 \\
YOLOv3 MobileNetV2 & 14.408 & 0.199 & 0.005 & 0.195 & 0.197 & 0.219 \\
YOLOX-l & 207.056 & 0.433 & 0.013 & 0.420 & 0.427 & 0.459 \\
YOLOX-s & 34.300 & 0.185 & 0.002 & 0.182 & 0.184 & 0.193 \\
YOLOX-tiny & 19.353 & 0.158 & 0.001 & 0.156 & 0.157 & 0.162 \\
YOLOX-x & 378.314 & 0.644 & 0.018 & 0.616 & 0.647 & 0.666 \\
\bottomrule
\end{tabular}
\label{tab:TitanX_time16_MMDet}
\end{table*}
\end{landscape}

\begin{landscape}
\begin{table*}[ht]
\centering
\caption{Time performances, in seconds, for MMDet presets with TitanX (batch size of 32).}
\begin{tabular}{lrrrrrr}
\toprule
Model & Size (MiB) & Time 32 mean & Time 32 std & Time 32 min & Time 32 median & Time 32 max \\
\midrule
Mask R-CNN R101 & 367.822 & - & - & - & - & - \\
Mask R-CNN R50 & 295.173 & - & - & - & - & - \\
Mask R-CNN X101 & 516.609 & - & - & - & - & - \\
R-CNN Faster R50 & 161.722 & - & - & - & - & - \\
Centernet R50 & 123.391 & 0.788 & 0.020 & 0.763 & 0.782 & 0.822 \\
Cond. DETR R50 & 166.023 & - & - & - & - & - \\
DAB DETR R50 & 166.989 & - & - & - & - & - \\
Deform. DETR R50 & 153.245 & - & - & - & - & - \\
Deform. DETR R50 (2 stages) & 157.421 & - & - & - & - & - \\
DETR R50 & 158.799 & - & - & - & - & - \\
DINO-4 R50 & 182.173 & - & - & - & - & - \\
DINO-5 Swin & 836.669 & - & - & - & - & - \\
Faster R-CNN Swin & 380.914 & 4.970 & 0.150 & 4.525 & 5.027 & 5.104 \\
Mask R-CNN Swin & 182.542 & 2.321 & 0.046 & 2.207 & 2.327 & 2.428 \\
Mask R-CNN Swin (Crop) & 182.542 & 2.288 & 0.046 & 2.186 & 2.282 & 2.401 \\
Retinanet & 76.619 & 1.335 & 0.041 & 1.279 & 1.322 & 1.396 \\
RTMDet-x & 362.298 & 3.787 & 0.022 & 3.749 & 3.784 & 3.890 \\
RTMDet-ins m & 105.383 & 0.897 & 0.027 & 0.857 & 0.896 & 0.949 \\
RTMDet-ins tiny & 21.579 & 0.516 & 0.010 & 0.503 & 0.514 & 0.549 \\
YOLOf R50 & 168.715 & 0.505 & 0.012 & 0.489 & 0.504 & 0.526 \\
YOLOv3 d53 & 236.519 & 0.695 & 0.017 & 0.670 & 0.702 & 0.718 \\
YOLOv3 MobileNetV2 & 14.408 & 0.397 & 0.009 & 0.388 & 0.393 & 0.424 \\
YOLOX-l & 207.056 & 0.857 & 0.027 & 0.822 & 0.869 & 0.893 \\
YOLOX-s & 34.300 & 0.370 & 0.008 & 0.361 & 0.367 & 0.402 \\
YOLOX-tiny & 19.353 & 0.314 & 0.002 & 0.312 & 0.314 & 0.322 \\
YOLOX-x & 378.314 & 1.288 & 0.028 & 1.226 & 1.303 & 1.318 \\
\bottomrule
\end{tabular}
\label{tab:TitanX_time32_MMDet}
\end{table*}
\end{landscape}

\begin{landscape}   
\begin{table*}[ht]
\centering
\caption{Time performances, in seconds, for MMDet presets with Quadro (batch size of 1).}
\begin{tabular}{lrrrrrr}
\toprule
Model & Size (MiB) & Time 1 mean & Time 1 std & Time 1 min & Time 1 median & Time 1 max \\
\midrule
Mask R-CNN R101 & 367.822 & 0.093 & 0.002 & 0.088 & 0.093 & 0.128 \\
Mask R-CNN R50 & 295.173 & 0.079 & 0.002 & 0.073 & 0.079 & 0.092 \\
Mask R-CNN X101 & 516.609 & 0.117 & 0.003 & 0.106 & 0.118 & 0.140 \\
R-CNN Faster R50 & 161.722 & 0.075 & 0.002 & 0.042 & 0.075 & 0.098 \\
Centernet R50 & 123.391 & 0.031 & 0.001 & 0.030 & 0.030 & 0.043 \\
Cond. DETR R50 & 166.023 & - & - & - & - & - \\
DAB DETR R50 & 166.989 & - & - & - & - & - \\
Deform. DETR R50 & 153.245 & - & - & - & - & - \\
Deform. DETR R50 (2 stages) & 157.421 & - & - & - & - & - \\
DETR R50 & 158.799 & - & - & - & - & - \\
DINO-4 R50 & 182.173 & - & - & - & - & - \\
DINO-5 Swin & 836.669 & - & - & - & - & - \\
Faster R-CNN Swin & 380.914 & 0.070 & 0.002 & 0.063 & 0.070 & 0.082 \\
Mask R-CNN Swin & 182.542 & 0.040 & 0.004 & 0.031 & 0.039 & 0.052 \\
Mask R-CNN Swin (Crop) & 182.542 & 0.034 & 0.002 & 0.030 & 0.034 & 0.056 \\
Retinanet & 76.619 & 0.045 & 0.001 & 0.043 & 0.045 & 0.084 \\
RTMDet-x & 362.298 & 0.093 & 0.001 & 0.088 & 0.093 & 0.107 \\
RTMDet-ins m & 105.383 & 0.027 & 0.001 & 0.026 & 0.027 & 0.037 \\
RTMDet-ins tiny & 21.579 & 0.022 & 0.001 & 0.021 & 0.022 & 0.043 \\
YOLOf R50 & 168.715 & 0.020 & 0.000 & 0.019 & 0.020 & 0.024 \\
YOLOv3 d53 & 236.519 & 0.023 & 0.001 & 0.021 & 0.022 & 0.033 \\
YOLOv3 MobileNetV2 & 14.408 & 0.014 & 0.000 & 0.014 & 0.014 & 0.023 \\
YOLOX-l & 207.056 & 0.025 & 0.001 & 0.024 & 0.025 & 0.038 \\
YOLOX-s & 34.300 & 0.018 & 0.001 & 0.018 & 0.018 & 0.027 \\
YOLOX-tiny & 19.353 & 0.018 & 0.001 & 0.018 & 0.018 & 0.037 \\
YOLOX-x & 378.314 & 0.042 & 0.001 & 0.037 & 0.042 & 0.049 \\
\bottomrule
\end{tabular}
\label{tab:Quadro_time1_MMDet}
\end{table*}
\end{landscape}

\begin{landscape}   
\begin{table*}[ht]
\centering
\caption{Time performances, in seconds, for MMDet presets with Quadro (batch size of 16).}
\begin{tabular}{lrrrrrr}
\toprule
Model & Size (MiB) & Time 16 mean & Time 16 std & Time 16 min & Time 16 median & Time 16 max \\
\midrule
Mask R-CNN R101 & 367.822 & 1.133 & 0.014 & 1.105 & 1.132 & 1.215 \\
Mask R-CNN R50 & 295.173 & 0.971 & 0.021 & 0.909 & 0.973 & 1.049 \\
Mask R-CNN X101 & 516.609 & 1.495 & 0.018 & 1.451 & 1.495 & 1.555 \\
R-CNN Faster R50 & 161.722 & 1.030 & 0.022 & 0.964 & 1.032 & 1.072 \\
Centernet R50 & 123.391 & 0.268 & 0.002 & 0.265 & 0.268 & 0.276 \\
Cond. DETR R50 & 166.023 & - & - & - & - & - \\
DAB DETR R50 & 166.989 & - & - & - & - & - \\
Deform. DETR R50 & 153.245 & - & - & - & - & - \\
Deform. DETR R50 (2 stages) & 157.421 & - & - & - & - & - \\
DETR R50 & 158.799 & - & - & - & - & - \\
DINO-4 R50 & 182.173 & - & - & - & - & - \\
DINO-5 Swin & 836.669 & - & - & - & - & - \\
Faster R-CNN Swin & 380.914 & 0.852 & 0.006 & 0.832 & 0.852 & 0.864 \\
Mask R-CNN Swin & 182.542 & 0.467 & 0.007 & 0.452 & 0.466 & 0.490 \\
Mask R-CNN Swin (Crop) & 182.542 & 0.453 & 0.006 & 0.439 & 0.452 & 0.478 \\
Retinanet & 76.619 & 0.396 & 0.003 & 0.389 & 0.396 & 0.405 \\
RTMDet-x & 362.298 & 1.381 & 0.008 & 1.361 & 1.381 & 1.417 \\
RTMDet-ins m & 105.383 & 0.276 & 0.002 & 0.272 & 0.276 & 0.281 \\
RTMDet-ins tiny & 21.579 & 0.160 & 0.001 & 0.157 & 0.160 & 0.169 \\
YOLOf R50 & 168.715 & 0.183 & 0.001 & 0.181 & 0.183 & 0.186 \\
YOLOv3 d53 & 236.519 & 0.242 & 0.001 & 0.240 & 0.242 & 0.244 \\
YOLOv3 MobileNetV2 & 14.408 & 0.117 & 0.001 & 0.116 & 0.117 & 0.124 \\
YOLOX-l & 207.056 & 0.307 & 0.002 & 0.304 & 0.307 & 0.312 \\
YOLOX-s & 34.300 & 0.138 & 0.001 & 0.136 & 0.138 & 0.143 \\
YOLOX-tiny & 19.353 & 0.121 & 0.001 & 0.119 & 0.121 & 0.124 \\
YOLOX-x & 378.314 & 0.493 & 0.003 & 0.485 & 0.493 & 0.505 \\
\bottomrule
\end{tabular}
\label{tab:Quadro_time16_MMDet}
\end{table*}
\end{landscape}

\begin{landscape}   
\begin{table*}[ht]
\centering
\caption{Time performances, in seconds, for MMDet presets with Quadro (batch size of 32).}
\begin{tabular}{lrrrrrr}
\toprule
Model & Size (MiB) & Time 32 mean & Time 32 std & Time 32 min & Time 32 median & Time 32 max \\
\midrule
Mask R-CNN R101 & 367.822 & 2.382 & 0.022 & 2.304 & 2.382 & 2.444 \\
Mask R-CNN R50 & 295.173 & 1.906 & 0.046 & 1.780 & 1.907 & 2.008 \\
Mask R-CNN X101 & 516.609 & 3.112 & 0.028 & 3.038 & 3.112 & 3.181 \\
R-CNN Faster R50 & 161.722 & 1.999 & 0.042 & 1.865 & 1.999 & 2.099 \\
Centernet R50 & 123.391 & 0.551 & 0.002 & 0.547 & 0.551 & 0.558 \\
Cond. DETR R50 & 166.023 & - & - & - & - & - \\
DAB DETR R50 & 166.989 & - & - & - & - & - \\
Deform. DETR R50 & 153.245 & - & - & - & - & - \\
Deform. DETR R50 (2 stages) & 157.421 & - & - & - & - & - \\
DETR R50 & 158.799 & - & - & - & - & - \\
DINO-4 R50 & 182.173 & - & - & - & - & - \\
DINO-5 Swin & 836.669 & - & - & - & - & - \\
Faster R-CNN Swin & 380.914 & 1.583 & 0.019 & 1.564 & 1.578 & 1.674 \\
Mask R-CNN Swin & 182.542 & 0.894 & 0.007 & 0.879 & 0.894 & 0.914 \\
Mask R-CNN Swin (Crop) & 182.542 & 0.916 & 0.006 & 0.899 & 0.916 & 0.933 \\
Retinanet & 76.619 & 0.787 & 0.003 & 0.779 & 0.787 & 0.796 \\
RTMDet-x & 362.298 & 2.767 & 0.125 & 2.684 & 2.714 & 3.275 \\
RTMDet-ins m & 105.383 & 0.544 & 0.003 & 0.538 & 0.544 & 0.551 \\
RTMDet-ins tiny & 21.579 & 0.320 & 0.002 & 0.315 & 0.320 & 0.336 \\
YOLOf R50 & 168.715 & 0.359 & 0.002 & 0.357 & 0.359 & 0.375 \\
YOLOv3 d53 & 236.519 & 0.482 & 0.001 & 0.479 & 0.481 & 0.489 \\
YOLOv3 MobileNetV2 & 14.408 & 0.235 & 0.001 & 0.234 & 0.235 & 0.239 \\
YOLOX-l & 207.056 & 0.606 & 0.002 & 0.602 & 0.606 & 0.613 \\
YOLOX-s & 34.300 & 0.275 & 0.002 & 0.273 & 0.275 & 0.290 \\
YOLOX-tiny & 19.353 & 0.242 & 0.001 & 0.240 & 0.242 & 0.245 \\
YOLOX-x & 378.314 & 0.926 & 0.004 & 0.916 & 0.925 & 0.935 \\
\bottomrule
\end{tabular}
\label{tab:Quadro_time32_MMDet}
\end{table*}
\end{landscape}

\begin{landscape}
\begin{table*}[ht]
\centering
\caption{Time performances, in seconds, for MMDet presets with A6000 (batch size of 1).}
\begin{tabular}{lrrrrrr}
\toprule
Model & Size (MiB) & Time 1 mean & Time 1 std & Time 1 min & Time 1 median & Time 1 max \\
\midrule
Mask R-CNN R101 & 367.822 & 0.068 & 0.013 & 0.056 & 0.064 & 0.194 \\
Mask R-CNN R50 & 295.173 & 0.051 & 0.003 & 0.046 & 0.050 & 0.072 \\
Mask R-CNN X100 & 516.609 & 0.071 & 0.012 & 0.060 & 0.067 & 0.213 \\
R-CNN Faster R50 & 161.722 & 0.068 & 0.026 & 0.031 & 0.068 & 0.252 \\
Centernet R50 & 123.391 & 0.033 & 0.009 & 0.026 & 0.030 & 0.143 \\
Cond. DETR R50 & 166.023 & - & - & - & - & - \\
DAB DETR R50 & 166.989 & - & - & - & - & - \\
Deform. DETR R50 & 153.245 & - & - & - & - & - \\
Deform. DETR R50 (2 stages) & 157.421 & - & - & - & - & - \\
DETR R50 & 158.799 & - & - & - & - & - \\
DINO-4 R50 & 182.173 & - & - & - & - & - \\
DINO-5 Swin & 836.669 & - & - & - & - & - \\
Faster R-CNN Swin & 380.914 & 0.040 & 0.009 & 0.032 & 0.037 & 0.121 \\
Mask R-CNN Swin & 182.542 & 0.033 & 0.011 & 0.023 & 0.029 & 0.138 \\
Mask R-CNN Swin (Crop) & 182.542 & 0.028 & 0.005 & 0.023 & 0.027 & 0.076 \\
Retinanet & 76.619 & 0.045 & 0.009 & 0.037 & 0.042 & 0.145 \\
RTMDet-x & 362.298 & 0.092 & 0.021 & 0.074 & 0.084 & 0.197 \\
RTMDet-ins m & 105.383 & 0.030 & 0.006 & 0.023 & 0.027 & 0.140 \\
RTMDet-ins tiny & 21.579 & 0.024 & 0.005 & 0.018 & 0.022 & 0.071 \\
YOLOf R50 & 168.715 & 0.021 & 0.005 & 0.016 & 0.019 & 0.089 \\
YOLOv3 d53 & 236.519 & 0.015 & 0.002 & 0.012 & 0.015 & 0.051 \\
YOLOv3 MobileNetV2 & 14.408 & 0.015 & 0.004 & 0.011 & 0.014 & 0.112 \\
YOLOX-l & 207.056 & 0.028 & 0.006 & 0.021 & 0.025 & 0.076 \\
YOLOX-s & 34.300 & 0.018 & 0.003 & 0.015 & 0.018 & 0.059 \\
YOLOX-tiny & 19.353 & 0.019 & 0.003 & 0.015 & 0.018 & 0.051 \\
YOLOX-x & 378.314 & 0.026 & 0.002 & 0.023 & 0.026 & 0.042 \\
\bottomrule
\end{tabular}
\label{tab:A6000_time1_MMDet}
\end{table*}
\end{landscape}

\begin{landscape}
\begin{table*}[ht]
\centering
\caption{Time performances, in seconds, for MMDet presets with A6000 (batch size of 16).}
\begin{tabular}{lrrrrrr}
\toprule
Model & Size (MiB) & Time 16 mean & Time 16 std & Time 16 min & Time 16 median & Time 16 max \\
\midrule
Mask R-CNN R101 & 367.822 & 0.644 & 0.008 & 0.629 & 0.642 & 0.673 \\
Mask R-CNN R50 & 295.173 & 0.679 & 0.013 & 0.638 & 0.678 & 0.709 \\
Mask R-CNN X100 & 516.609 & 0.815 & 0.007 & 0.799 & 0.814 & 0.847 \\
R-CNN Faster R50 & 161.722 & 0.904 & 0.042 & 0.791 & 0.900 & 1.104 \\
Centernet R50 & 123.391 & 0.194 & 0.010 & 0.182 & 0.191 & 0.241 \\
Cond. DETR R50 & 166.023 & - & - & - & - & - \\
DAB DETR R50 & 166.989 & - & - & - & - & - \\
Deform. DETR R50 & 153.245 & - & - & - & - & - \\
Deform. DETR R50 (2 stages) & 157.421 & - & - & - & - & - \\
DETR R50 & 158.799 & - & - & - & - & - \\
DINO-4 R50 & 182.173 & - & - & - & - & - \\
DINO-5 Swin & 836.669 & - & - & - & - & - \\
Faster R-CNN Swin & 380.914 & 0.521 & 0.009 & 0.508 & 0.519 & 0.560 \\
Mask R-CNN Swin & 182.542 & 0.351 & 0.013 & 0.332 & 0.349 & 0.422 \\
Mask R-CNN Swin (Crop) & 182.542 & 0.346 & 0.008 & 0.332 & 0.345 & 0.397 \\
Retinanet & 76.619 & 0.293 & 0.005 & 0.286 & 0.292 & 0.322 \\
RTMDet-x & 362.298 & 1.206 & 0.031 & 1.127 & 1.205 & 1.303 \\
RTMDet-ins m & 105.383 & 0.226 & 0.017 & 0.204 & 0.219 & 0.280 \\
RTMDet-ins tiny & 21.579 & 0.158 & 0.021 & 0.134 & 0.148 & 0.239 \\
YOLOf R50 & 168.715 & 0.133 & 0.011 & 0.122 & 0.128 & 0.192 \\
YOLOv3 d53 & 236.519 & 0.164 & 0.007 & 0.157 & 0.162 & 0.210 \\
YOLOv3 MobileNetV2 & 14.408 & 0.098 & 0.012 & 0.088 & 0.093 & 0.154 \\
YOLOX-l & 207.056 & 0.222 & 0.006 & 0.214 & 0.220 & 0.247 \\
YOLOX-s & 34.300 & 0.128 & 0.016 & 0.116 & 0.121 & 0.190 \\
YOLOX-tiny & 19.353 & 0.117 & 0.015 & 0.105 & 0.112 & 0.204 \\
YOLOX-x & 378.314 & 0.304 & 0.003 & 0.300 & 0.302 & 0.316 \\
\bottomrule
\end{tabular}
\label{tab:A6000_time16_MMDet}
\end{table*}
\end{landscape}

\begin{landscape}
\begin{table*}[ht]
\centering
\caption{Time performances, in seconds, for MMDet presets with A6000 (batch size of 32).}
\begin{tabular}{lrrrrrr}
\toprule
Model & Size (MiB) & Time 32 mean & Time 32 std & Time 32 min & Time 32 median & Time 32 max \\
\midrule
Mask R-CNN R101 & 367.822 & 1.341 & 0.012 & 1.313 & 1.339 & 1.383 \\
Mask R-CNN R50 & 295.173 & 1.127 & 0.021 & 1.069 & 1.131 & 1.168 \\
Mask R-CNN X100 & 516.609 & 1.848 & 0.016 & 1.817 & 1.845 & 1.933 \\
R-CNN Faster R50 & 161.722 & 1.639 & 0.046 & 1.558 & 1.633 & 1.783 \\
Centernet R50 & 123.391 & 0.405 & 0.012 & 0.388 & 0.404 & 0.448 \\
Cond. DETR R50 & 166.023 & - & - & - & - & - \\
DAB DETR R50 & 166.989 & - & - & - & - & - \\
Deform. DETR R50 & 153.245 & - & - & - & - & - \\
Deform. DETR R50 (2 stages) & 157.421 & - & - & - & - & - \\
DETR R50 & 158.799 & - & - & - & - & - \\
DINO-4 R50 & 182.173 & - & - & - & - & - \\
DINO-5 Swin & 836.669 & - & - & - & - & - \\
Faster R-CNN Swin & 380.914 & 0.915 & 0.011 & 0.898 & 0.913 & 0.960 \\
Mask R-CNN Swin & 182.542 & 0.698 & 0.014 & 0.674 & 0.696 & 0.742 \\
Mask R-CNN Swin (Crop) & 182.542 & 0.753 & 0.019 & 0.714 & 0.752 & 0.822 \\
Retinanet & 76.619 & 0.574 & 0.007 & 0.564 & 0.572 & 0.611 \\
RTMDet-x & 362.298 & 2.351 & 0.071 & 2.226 & 2.347 & 2.665 \\
RTMDet-ins m & 105.383 & 0.410 & 0.013 & 0.394 & 0.405 & 0.448 \\
RTMDet-ins tiny & 21.579 & 0.294 & 0.016 & 0.270 & 0.289 & 0.361 \\
YOLOf R50 & 168.715 & 0.244 & 0.008 & 0.234 & 0.243 & 0.270 \\
YOLOv3 d53 & 236.519 & 0.316 & 0.009 & 0.303 & 0.313 & 0.347 \\
YOLOv3 MobileNetV2 & 14.408 & 0.196 & 0.016 & 0.178 & 0.188 & 0.270 \\
YOLOX-l & 207.056 & 0.426 & 0.007 & 0.416 & 0.423 & 0.450 \\
YOLOX-s & 34.300 & 0.245 & 0.013 & 0.229 & 0.242 & 0.279 \\
YOLOX-tiny & 19.353 & 0.229 & 0.017 & 0.211 & 0.225 & 0.285 \\
YOLOX-x & 378.314 & 0.595 & 0.002 & 0.592 & 0.595 & 0.608 \\
\bottomrule
\end{tabular}
\label{tab:A6000_time32_MMDet}
\end{table*}
\end{landscape}

\begin{landscape}
\begin{table*}[ht]
\centering
\caption{Time performances, in seconds, for MMDet presets with RTX-3080Ti (batch size of 1).}
\begin{tabular}{lrrrrrr}
\toprule
Model & Size (MiB) & Time 1 mean & Time 1 std & Time 1 min & Time 1 median & Time 1 max \\
\midrule
Mask R-CNN R101 & 367.822 & 0.049 & 0.002 & 0.046 & 0.049 & 0.095 \\
Mask R-CNN R50 & 295.173 & 0.047 & 0.001 & 0.044 & 0.047 & 0.061 \\
Mask R-CNN X101 & 516.609 & 0.066 & 0.002 & 0.061 & 0.065 & 0.092 \\
R-CNN Faster R50 & 161.722 & 0.038 & 0.001 & 0.036 & 0.038 & 0.070 \\
Centernet R50 & 123.391 & 0.015 & 0.001 & 0.015 & 0.015 & 0.034 \\
Cond. DETR R50 & 166.023 & - & - & - & - & - \\
DAB DETR R50 & 166.989 & - & - & - & - & - \\
Deform. DETR R50 & 153.245 & - & - & - & - & - \\
Deform. DETR R50 (2 stages) & 157.421 & - & - & - & - & - \\
DETR R50 & 158.799 & - & - & - & - & - \\
DINO-4 R50 & 182.173 & - & - & - & - & - \\
DINO-5 Swin & 836.669 & - & - & - & - & - \\
Faster R-CNN Swin & 380.914 & 0.036 & 0.002 & 0.033 & 0.036 & 0.044 \\
Mask R-CNN Swin & 182.542 & 0.026 & 0.002 & 0.021 & 0.025 & 0.053 \\
Mask R-CNN Swin (Crop) & 182.542 & 0.028 & 0.006 & 0.021 & 0.026 & 0.079 \\
Retinanet & 76.619 & 0.024 & 0.001 & 0.023 & 0.024 & 0.037 \\
RTMDet-x & 362.298 & 0.063 & 0.003 & 0.059 & 0.062 & 0.122 \\
RTMDet-ins m & 105.383 & 0.014 & 0.001 & 0.013 & 0.014 & 0.039 \\
RTMDet-ins tiny & 21.579 & 0.011 & 0.001 & 0.010 & 0.011 & 0.024 \\
YOLOf R50 & 168.715 & 0.010 & 0.001 & 0.010 & 0.010 & 0.027 \\
YOLOv3 d53 & 236.519 & 0.014 & 0.000 & 0.013 & 0.013 & 0.022 \\
YOLOv3 MobileNetV2 & 14.408 & 0.006 & 0.001 & 0.006 & 0.006 & 0.018 \\
YOLOX-l & 207.056 & 0.017 & 0.001 & 0.016 & 0.017 & 0.027 \\
YOLOX-s & 34.300 & 0.009 & 0.000 & 0.009 & 0.009 & 0.019 \\
YOLOX-tiny & 19.353 & 0.009 & 0.000 & 0.009 & 0.009 & 0.023 \\
YOLOX-x & 378.314 & 0.024 & 0.002 & 0.023 & 0.024 & 0.047 \\
\bottomrule
\end{tabular}
\label{tab:RTX3080_time1_MMDet}
\end{table*}
\end{landscape}

\begin{landscape}
\begin{table*}[ht]
\centering
\caption{Time performances, in seconds, for MMDet presets with RTX-3080Ti (batch size of 16).}
\begin{tabular}{lrrrrrr}
\toprule
Model & Size (MiB) & Time 16 mean & Time 16 std & Time 16 min & Time 16 median & Time 16 max \\
\midrule
Mask R-CNN R101 & 367.822 & 0.690 & 0.009 & 0.665 & 0.690 & 0.767 \\
Mask R-CNN R50 & 295.173 & 0.721 & 0.046 & 0.649 & 0.712 & 1.007 \\
Mask R-CNN X101 & 516.609 & 0.860 & 0.103 & 0.784 & 0.825 & 1.321 \\
R-CNN Faster R50 & 161.722 & 0.524 & 0.011 & 0.490 & 0.524 & 0.597 \\
Centernet R50 & 123.391 & 0.190 & 0.003 & 0.188 & 0.190 & 0.219 \\
Cond. DETR R50 & 166.023 & - & - & - & - & - \\
DAB DETR R50 & 166.989 & - & - & - & - & - \\
Deform. DETR R50 & 153.245 & - & - & - & - & - \\
Deform. DETR R50 (2 stages) & 157.421 & - & - & - & - & - \\
DETR R50 & 158.799 & - & - & - & - & - \\
DINO-4 R50 & 182.173 & - & - & - & - & - \\
DINO-5 Swin & 836.669 & - & - & - & - & - \\
Faster R-CNN Swin & 380.914 & 0.451 & 0.016 & 0.426 & 0.446 & 0.513 \\
Mask R-CNN Swin & 182.542 & 0.343 & 0.025 & 0.313 & 0.336 & 0.498 \\
Mask R-CNN Swin (Crop) & 182.542 & 0.390 & 0.027 & 0.350 & 0.384 & 0.504 \\
Retinanet & 76.619 & 0.279 & 0.010 & 0.267 & 0.275 & 0.324 \\
RTMDet-x & 362.298 & 0.985 & 0.049 & 0.925 & 0.980 & 1.109 \\
RTMDet-ins m & 105.383 & 0.188 & 0.002 & 0.185 & 0.188 & 0.202 \\
RTMDet-ins tiny & 21.579 & 0.109 & 0.002 & 0.105 & 0.109 & 0.124 \\
YOLOf R50 & 168.715 & 0.114 & 0.003 & 0.111 & 0.113 & 0.137 \\
YOLOv3 d53 & 236.519 & 0.167 & 0.003 & 0.164 & 0.166 & 0.208 \\
YOLOv3 MobileNetV2 & 14.408 & 0.079 & 0.002 & 0.077 & 0.079 & 0.102 \\
YOLOX-l & 207.056 & 0.216 & 0.006 & 0.212 & 0.213 & 0.261 \\
YOLOX-s & 34.300 & 0.105 & 0.001 & 0.104 & 0.105 & 0.126 \\
YOLOX-tiny & 19.353 & 0.095 & 0.001 & 0.094 & 0.095 & 0.108 \\
YOLOX-x & 378.314 & 0.341 & 0.015 & 0.329 & 0.333 & 0.404 \\
\bottomrule
\end{tabular}
\label{tab:RTX3080_time16_MMDet}
\end{table*}
\end{landscape}

\begin{landscape}
\begin{table*}[ht]
\centering
\caption{Time performances, in seconds, for MMDet presets with RTX-3080Ti (batch size of 32).}
\begin{tabular}{lrrrrrr}
\toprule
Model & Size (MiB) & Time 32 mean & Time 32 std & Time 32 min & Time 32 median & Time 32 max \\
\midrule
Mask R-CNN R101 & 367.822 & - & - & - & - & - \\
Mask R-CNN R50 & 295.173 & - & - & - & - & - \\
Mask R-CNN X101 & 516.609 & - & - & - & - & - \\
R-CNN Faster R50 & 161.722 & - & - & - & - & - \\
Centernet R50 & 123.391 & 0.355 & 0.004 & 0.350 & 0.354 & 0.376 \\
Cond. DETR R50 & 166.023 & - & - & - & - & - \\
DAB DETR R50 & 166.989 & - & - & - & - & - \\
Deform. DETR R50 & 153.245 & - & - & - & - & - \\
Deform. DETR R50 (2 stages) & 157.421 & - & - & - & - & - \\
DETR R50 & 158.799 & - & - & - & - & - \\
DINO-4 R50 & 182.173 & - & - & - & - & - \\
DINO-5 Swin & 836.669 & - & - & - & - & - \\
Faster R-CNN Swin & 380.914 & 0.994 & 0.015 & 0.955 & 0.994 & 1.038 \\
Mask R-CNN Swin & 182.542 & 0.666 & 0.046 & 0.613 & 0.652 & 0.879 \\
Mask R-CNN Swin (Crop) & 182.542 & 0.690 & 0.055 & 0.636 & 0.675 & 0.941 \\
Retinanet & 76.619 & 0.587 & 0.037 & 0.528 & 0.583 & 0.654 \\
RTMDet-x & 362.298 & 1.984 & 0.062 & 1.886 & 1.969 & 2.184 \\
RTMDet-ins m & 105.383 & 0.380 & 0.004 & 0.375 & 0.378 & 0.406 \\
RTMDet-ins tiny & 21.579 & 0.214 & 0.002 & 0.210 & 0.214 & 0.237 \\
YOLOf R50 & 168.715 & 0.223 & 0.001 & 0.222 & 0.223 & 0.227 \\
YOLOv3 d53 & 236.519 & 0.336 & 0.008 & 0.324 & 0.337 & 0.359 \\
YOLOv3 MobileNetV2 & 14.408 & 0.156 & 0.005 & 0.154 & 0.156 & 0.212 \\
YOLOX-l & 207.056 & 0.419 & 0.002 & 0.417 & 0.418 & 0.441 \\
YOLOX-s & 34.300 & 0.207 & 0.002 & 0.206 & 0.207 & 0.217 \\
YOLOX-tiny & 19.353 & 0.195 & 0.007 & 0.188 & 0.189 & 0.217 \\
YOLOX-x & 378.314 & 0.678 & 0.028 & 0.649 & 0.667 & 0.777 \\
\bottomrule
\end{tabular}
\label{tab:RTX3080_time32_MMDet}
\end{table*}
\end{landscape}

\begin{landscape}
\begin{table*}[ht]
\centering
\caption{Time performances, in seconds, for MMDet presets with RTX-4090 (batch size of 1).}
\begin{tabular}{lrrrrrr}
\toprule
Model & Size (MiB) & Time 1 mean & Time 1 std & Time 1 min & Time 1 median & Time 1 max \\
\midrule
Mask R-CNN R101 & 367.822 & 0.032 & 0.002 & 0.029 & 0.031 & 0.049 \\
Mask R-CNN R50 & 295.173 & 0.027 & 0.001 & 0.025 & 0.026 & 0.039 \\
Mask R-CNN X100 & 516.609 & 0.030 & 0.001 & 0.028 & 0.030 & 0.045 \\
R-CNN Faster R50 & 161.722 & 0.029 & 0.002 & 0.017 & 0.029 & 0.057 \\
Centernet R50 & 123.391 & 0.011 & 0.001 & 0.010 & 0.011 & 0.023 \\
Cond. DETR R50 & 166.023 & - & - & - & - & - \\
DAB DETR R50 & 166.989 & - & - & - & - & - \\
Deformable DETR R50 & 153.245 & - & - & - & - & - \\
Deformable DETR R50 (2 stages) & 157.421 & - & - & - & - & - \\
DETR R50 & 158.799 & - & - & - & - & - \\
DINO-4 R50 & 182.173 & - & - & - & - & - \\
DINO-5 Swin & 836.669 & - & - & - & - & - \\
Faster R-CNN Swin & 380.914 & 0.016 & 0.001 & 0.015 & 0.016 & 0.030 \\
Mask R-CNN Swin & 182.542 & 0.014 & 0.001 & 0.012 & 0.013 & 0.025 \\
Mask R-CNN Swin (Crop) & 182.542 & 0.013 & 0.001 & 0.012 & 0.014 & 0.024 \\
Retinanet & 76.619 & 0.017 & 0.001 & 0.015 & 0.016 & 0.032 \\
RTMDet-x & 362.298 & 0.041 & 0.001 & 0.040 & 0.041 & 0.067 \\
RTMDet-ins m & 105.383 & 0.011 & 0.002 & 0.010 & 0.010 & 0.024 \\
RTMDet-ins tiny & 21.579 & 0.009 & 0.001 & 0.008 & 0.008 & 0.020 \\
YOLOf R50 & 168.715 & 0.007 & 0.001 & 0.007 & 0.007 & 0.016 \\
YOLOv3 d53 & 236.519 & 0.007 & 0.000 & 0.006 & 0.007 & 0.013 \\
YOLOv3 MobileNetV2 & 14.408 & 0.005 & 0.000 & 0.005 & 0.005 & 0.012 \\
YOLOX-l & 207.056 & 0.009 & 0.001 & 0.009 & 0.009 & 0.020 \\
YOLOX-s & 34.300 & 0.007 & 0.001 & 0.006 & 0.007 & 0.017 \\
YOLOX-tiny & 19.353 & 0.007 & 0.001 & 0.006 & 0.006 & 0.016 \\
YOLOX-x & 378.314 & 0.013 & 0.001 & 0.013 & 0.013 & 0.025 \\
\bottomrule
\end{tabular}
\label{tab:RTX-4090_time1_MMDet}
\end{table*}
\end{landscape}

\begin{landscape}
\begin{table*}[ht]
\centering
\caption{Time performances, in seconds, for MMDet presets with RTX-4090 (batch size of 16).}
\begin{tabular}{lrrrrrr}
\toprule
Model & Size (MiB) & Time 16 mean & Time 16 std & Time 16 min & Time 16 median & Time 16 max \\
\midrule
Mask R-CNN R101 & 367.822 & 0.409 & 0.014 & 0.389 & 0.403 & 0.448 \\
Mask R-CNN R50 & 295.173 & 0.370 & 0.010 & 0.349 & 0.367 & 0.405 \\
Mask R-CNN X100 & 516.609 & 0.478 & 0.012 & 0.466 & 0.473 & 0.531 \\
R-CNN Faster R50 & 161.722 & 0.398 & 0.007 & 0.379 & 0.400 & 0.416 \\
Centernet R50 & 123.391 & 0.112 & 0.002 & 0.111 & 0.111 & 0.122 \\
Cond. DETR R50 & 166.023 & - & - & - & - & - \\
DAB DETR R50 & 166.989 & - & - & - & - & - \\
Deformable DETR R50 & 153.245 & - & - & - & - & - \\
Deformable DETR R50 (2 stages) & 157.421 & - & - & - & - & - \\
DETR R50 & 158.799 & - & - & - & - & - \\
DINO-4 R50 & 182.173 & - & - & - & - & - \\
DINO-5 Swin & 836.669 & - & - & - & - & - \\
Faster R-CNN Swin & 380.914 & 0.307 & 0.010 & 0.286 & 0.305 & 0.338 \\
Mask R-CNN Swin & 182.542 & 0.190 & 0.004 & 0.180 & 0.189 & 0.205 \\
Mask R-CNN Swin (Crop) & 182.542 & 0.190 & 0.003 & 0.185 & 0.190 & 0.199 \\
Retinanet & 76.619 & 0.178 & 0.002 & 0.175 & 0.178 & 0.188 \\
RTMDet-x & 362.298 & 0.646 & 0.026 & 0.622 & 0.632 & 0.758 \\
RTMDet-ins m & 105.383 & 0.118 & 0.003 & 0.114 & 0.117 & 0.131 \\
RTMDet-ins tiny & 21.579 & 0.073 & 0.003 & 0.070 & 0.073 & 0.085 \\
YOLOf R50 & 168.715 & 0.080 & 0.003 & 0.078 & 0.078 & 0.091 \\
YOLOv3 d53 & 236.519 & 0.103 & 0.004 & 0.098 & 0.102 & 0.114 \\
YOLOv3 MobileNetV2 & 14.408 & 0.052 & 0.002 & 0.051 & 0.051 & 0.060 \\
YOLOX-l & 207.056 & 0.129 & 0.004 & 0.126 & 0.127 & 0.141 \\
YOLOX-s & 34.300 & 0.064 & 0.003 & 0.060 & 0.064 & 0.074 \\
YOLOX-tiny & 19.353 & 0.056 & 0.002 & 0.054 & 0.055 & 0.063 \\
YOLOX-x & 378.314 & 0.192 & 0.006 & 0.187 & 0.189 & 0.223 \\
\bottomrule
\end{tabular}
\label{tab:RTX-4090_time16_MMDet}
\end{table*}
\end{landscape}

\begin{landscape}
\begin{table*}[ht]
\centering
\caption{Time performances, in seconds, for MMDet presets with RTX-4090 (batch size of 32).}
\begin{tabular}{lrrrrrr}
\toprule
Model & Size (MiB) & Time 32 mean & Time 32 std & Time 32 min & Time 32 median & Time 32 max \\
\midrule
Mask R-CNN R101 & 367.822 & 0.909 & 0.028 & 0.864 & 0.906 & 0.987 \\
Mask R-CNN R50 & 295.173 & 0.702 & 0.021 & 0.678 & 0.697 & 0.854 \\
Mask R-CNN X100 & 516.609 & 1.020 & 0.032 & 0.980 & 1.007 & 1.096 \\
R-CNN Faster R50 & 161.722 & 0.823 & 0.008 & 0.809 & 0.821 & 0.901 \\
Centernet R50 & 123.391 & 0.241 & 0.002 & 0.240 & 0.241 & 0.251 \\
Cond. DETR R50 & 166.023 & - & - & - & - & - \\
DAB DETR R50 & 166.989 & - & - & - & - & - \\
Deformable DETR R50 & 153.245 & - & - & - & - & - \\
Deformable DETR R50 (2 stages) & 157.421 & - & - & - & - & - \\
DETR R50 & 158.799 & - & - & - & - & - \\
DINO-4 R50 & 182.173 & - & - & - & - & - \\
DINO-5 Swin & 836.669 & - & - & - & - & - \\
Faster R-CNN Swin & 380.914 & 0.577 & 0.020 & 0.557 & 0.567 & 0.632 \\
Mask R-CNN Swin & 182.542 & 0.403 & 0.007 & 0.389 & 0.402 & 0.425 \\
Mask R-CNN Swin (Crop) & 182.542 & 0.405 & 0.007 & 0.389 & 0.404 & 0.425 \\
Retinanet & 76.619 & 0.366 & 0.007 & 0.360 & 0.364 & 0.414 \\
RTMDet-x & 362.298 & 1.310 & 0.027 & 1.269 & 1.309 & 1.411 \\
RTMDet-ins m & 105.383 & 0.250 & 0.004 & 0.243 & 0.249 & 0.269 \\
RTMDet-ins tiny & 21.579 & 0.153 & 0.002 & 0.151 & 0.153 & 0.163 \\
YOLOf R50 & 168.715 & 0.162 & 0.004 & 0.158 & 0.160 & 0.175 \\
YOLOv3 d53 & 236.519 & 0.205 & 0.005 & 0.200 & 0.204 & 0.232 \\
YOLOv3 MobileNetV2 & 14.408 & 0.111 & 0.004 & 0.108 & 0.109 & 0.124 \\
YOLOX-l & 207.056 & 0.265 & 0.002 & 0.264 & 0.265 & 0.283 \\
YOLOX-s & 34.300 & 0.132 & 0.004 & 0.126 & 0.130 & 0.144 \\
YOLOX-tiny & 19.353 & 0.115 & 0.003 & 0.112 & 0.114 & 0.127 \\
YOLOX-x & 378.314 & 0.406 & 0.009 & 0.397 & 0.402 & 0.440 \\
\bottomrule
\end{tabular}
\label{tab:RTX-4090_time32_MMDet}
\end{table*}
\end{landscape}

\end{document}